\documentclass{article}
\usepackage{float}
\usepackage[final]{neurips_2026} 
\usepackage[utf8]{inputenc} 
\usepackage[T1]{fontenc}  
\usepackage{hyperref}      
\usepackage{url}           
\usepackage{booktabs}      
\usepackage{amsfonts}    
\usepackage{amsmath}       
\usepackage{nicefrac}     
\usepackage{microtype}   
\usepackage{xcolor}      
\usepackage{graphicx}
\usepackage{url}

\makeatletter
\renewcommand{\@noticestring}{%
  This paper is accepted at \textbf{AI4Science} and \textbf{NewinML} Workshops (NeurIPS 2026).%
}
\makeatother
 
\title{Profile-Based Auditing of Domestic Violence Survey Selection Across South Asia}

\author{%
  Mohammad Abu Obaida Mullick\footnotemark[1]\\
  Institute for Advanced Research \\
  United International University \\
  Dhaka, Bangladesh \\
  \And
  Rezaur Rahman Ratul\footnotemark[1]\\
  Dept of Computer Science \& Engineering\\
  United International University \\
  Dhaka, Bangladesh \\
  \And
  Mymuna Nourin \\
  Dept of Computer Science \& Engineering\\
  United International University \\
  Dhaka, Bangladesh \\
  \And
  Md.~Motaharul Islam \\
  Dept of Computer Science \& Engineering\\
  United International University \\
  Dhaka, Bangladesh \\
\And
  Correspondence: mmullick212057@bscse.uiu.ac.bd
}

\begin{document}

\maketitle

\footnotetext[1]{%
  \begin{minipage}[t]{\linewidth}
    Equal contribution (co-first authors).\par
    Code: \href{https://github.com/maomullick/Audit-DomVio-SurveyData-SouthAsia}{https://github.com/maomullick/Audit-DomVio-SurveyData-SouthAsia}
  \end{minipage}%
}

\begin{abstract}
  Domestic violence (DV) data in the Demographic and Health Surveys (DHS) are collected through a specialized module whose household-level sampling and privacy requirements create a distinct observation process. We ask whether demographic, socioeconomic, and household characteristics form reproducible population profiles and whether these profiles contain information about selection into the DV module across South Asia. Using DHS Individual Recode data for 796,374 women aged 15--49 across Afghanistan, India, Myanmar, Nepal, and Pakistan, we harmonize 61 numeric and 8 categorical features, reduce the profile space with PCA, and identify six interpretable profiles. We then evaluate 22 machine-learning models under primary-sampling-unit-aware holdout validation. The best model, HistGradientBoosting, achieves ROC-AUC 0.9167 and PR-AUC 0.9612, versus 0.8475 and 0.9203 for a design-only baseline. Adding woman-level characteristics yields the largest incremental gain, while explicit interaction terms add little out-of-sample performance. Country-specific analyses reveal substantial heterogeneity, and leave-one-country-out evaluation shows that transferability of the incremental profile signal is limited. These results demonstrate that interpretable ML can audit differential coverage in sensitive survey modules beyond the formal sampling design.
\end{abstract}

\section{Introduction}
Domestic violence remains a major concern across South Asia, where women's experiences and household circumstances are shaped by interconnected demographic, socioeconomic, and family characteristics. Large-scale household surveys such as the Demographic and Health Surveys (DHS) provide an important empirical foundation for studying these populations. Yet domestic-violence information in DHS is collected through a specialized module whose sampling and fieldwork procedures differ from those of the standard women's questionnaire. In particular, selection for the domestic-violence module is implemented within households, and the sensitive interview is subject to privacy requirements \citep{dhs_dv_module,dhs_guide8,who2001}. Consequently, the women for whom domestic-violence information is observed form a selected population rather than a simple copy of the broader female DHS sample.

This distinction motivates a question that is largely separate from estimating the prevalence of domestic violence itself: \emph{what types of women and household contexts are represented in the observed domestic-violence module, and does that representation vary systematically across countries?} Conventional analyses typically examine demographic and socioeconomic characteristics individually, or treat country as a contextual covariate. Such approaches can miss multidimensional configurations in which characteristics co-occur. For example, age, education, marital status, wealth, residence, fertility, household composition, and women's agency may jointly define recurring population profiles even when no single variable is sufficient to describe them. We therefore formulate the problem as a profile-learning and selection-analysis task. Our main research question is:\\\\
\emph{\textbf{What distinct demographic and socioeconomic profiles of women and households can be identified across South Asian countries, how do these profiles interact, and to what extent are their combinations associated with household selection for the DHS domestic-violence module?}}\\\\
The question has four linked components. First, we identify recurring demographic and socioeconomic profiles in the female DHS population. Second, we characterize how those profiles are distributed across countries. Third, we test whether the underlying demographic, socioeconomic, household, agency, and normative characteristics contain information about the realized DV-module selection indicator beyond the household sampling design and country. Fourth, we examine whether the learned relationship transfers across countries.

Our analysis uses harmonized DHS Individual Recode data for Afghanistan, India, Myanmar, Nepal, and Pakistan. The resulting modeling universe contains 796,374 women aged 15--49. We construct a multidimensional feature space spanning household design variables, country, woman-level characteristics, household characteristics, agency measures, and normative indicators. Profile construction is performed without using the DV-selection variables themselves: the profile representation is learned from 51 numeric and 5 categorical characteristics, with design variables deliberately excluded from the profile space. Principal component analysis retains 19 components explaining 81.2\% of variance, after which a six-profile reporting solution is fitted and assigned to the complete modeling population.

We then approach DV-module selection as a supervised prediction problem, but with an important methodological qualification. The target $V044$ is a sampling and fieldwork indicator, not an indicator of experiencing domestic violence. The first supervised model is therefore not interpreted as a ''DV risk'' model. Instead, we ask whether the demographic--socioeconomic feature space contains predictive information beyond what the DHS design itself already implies. This distinction is central: a strong design-only model is expected because the number of eligible women in the household directly determines the within-household selection probability.

The study makes three contributions. First, it provides a multivariate description of demographic and socioeconomic profiles in a large South Asian DHS population. Second, it quantifies how much profile information adds to the prediction of DV-module selection beyond design and country variables. Third, it tests whether this incremental signal is stable across national contexts rather than assuming that a pooled relationship transfers uniformly.

The resulting perspective is deliberately about \emph{measurement and selection}, not about identifying the causes of domestic violence. By separating the broader population, module selection, and subsequent interview completion, the analysis provides a framework for understanding how sensitive-survey observations are assembled and how their representation can vary across countries.

\section{Data and Study Design}

\subsection{DHS Data Sources and Analytic Population}

The analysis uses DHS recode files organized around three record types. Household Recode (HR) files provide household-level information; Household Member Recode (PR) files represent listed household members and support construction of roster-based female study frames; and Individual Recode (IR) files contain one record per interviewed woman aged 15--49 and provide the woman-level variables used in the principal machine-learning analysis. The study's recode map and archive inventory document the survey stems, variables, and record structures in detail (Appendix~\ref{app:dhs_architecture}).

For the machine-learning analysis, we restrict the Individual Recode data to the five countries for which the DV-module selection indicator $V044$ is available and analyzable: Afghanistan, India, Myanmar, Nepal, and Pakistan. The resulting selection-analysis universe contains 796,374 women aged 15--49. Bangladesh and Sri Lanka are present in the broader survey materials but are not included in the $V044$ modeling universe because the supplied IR data do not provide the required domestic-violence selection information. The complete survey inventory and country-specific sample flows are reported in Appendix~\ref{app:dhs_architecture}.

Table~\ref{tab:main_sample} summarizes the principal modeling population and the realized DV-module selection and interview outcomes. The large cross-country differences in selection and completion rates provide the context for the subsequent profile and ML analysis.

\begin{table}[!ht]
\centering
\caption{Study population and realized DV-module selection in the five-country modeling universe.}
\label{tab:main_sample}
\small
\begin{tabular}{lrrrrr}
\toprule
Country & Women & Selected & Selected (\%) & Interviewed & Interviewed/Selected (\%) \\
\midrule
Afghanistan & 29,461 & 22,132 & 75.12 & 21,324 & 96.35 \\
India      & 724,115 & 494,830 & 68.34 & 72,320 & 14.62 \\
Myanmar    & 12,885 & 4,563 & 35.41 & 4,530 & 99.28 \\
Nepal      & 14,845 & 5,192 & 34.97 & 5,178 & 99.73 \\
Pakistan   & 15,068 & 4,216 & 27.98 & 4,085 & 96.89 \\
\midrule
Total      & 796,374 & 530,933 & 66.67 & 107,437 & 20.24 \\
\bottomrule
\end{tabular}

\vspace{1mm}
\begin{minipage}{0.96\linewidth}
\footnotesize
\textit{Note.} ''Selected'' corresponds to the study target $y_{\mathrm{selected}}$, defined from $V044\in\{1,2,3\}$. ''Interviewed'' corresponds to $V044=1$. Country totals are the modeling-dataset counts produced by the study. Detailed country-specific denominator flows and $V044$ distributions are given in Appendix~\ref{app:dv_analysis}.
\end{minipage}
\end{table}

\subsection{Selection Target and Design Structure}

The principal supervised-learning target is
\begin{equation}
y_i =
\mathbf{1}{V044_i\in{1,2,3}}
\end{equation}

where $y_i=1$ denotes that woman $i$ was selected for the DHS domestic-violence module, and $y_i=0$ denotes that she was not selected. Within the selected group, the study defines $y_{\mathrm{completed}}=1$ when $V044=1$; $V044=2$ indicates that privacy was not possible, and $V044=3$ indicates that the selected woman was not interviewed for another reason. This second process is treated as a separate completion analysis and is reported in Appendix~\ref{app:completion}.

The central design relationship is represented by the number of eligible women in the household, $V138$:
\begin{equation}
P(\text{selected})=P(\text{household in the DV sub-sample})
\times
\frac{1}{V138}.
\end{equation}

The mean number of eligible women in the modeling universe is 1.78. This design relationship is crucial for interpreting every predictive result. A high AUC by itself is not evidence that demographic characteristics distort module selection, because $V138$ already contains strong information about the outcome. The meaningful quantity for the research question is the \emph{incremental predictive information beyond design and country}. The study therefore implements an explicit design-versus-profile feature ladder (Appendix~\ref{app:selection_models}).

\subsection{Feature Construction}

The modeling matrix combines six substantive feature blocks plus an explicit interaction block. The design block contains the number of eligible women, its reciprocal, indicators for a single eligible woman, capped eligible-women categories, household size, number of children under five, and crowding. Country context is represented through survey year in the pooled feature matrix.

The woman-level block includes age, educational attainment, marital and union status, age at cohabitation, child-marriage indicators, fertility and birth-timing variables, contraceptive-use variables, and partner/spousal characteristics. The household block includes urban residence, wealth, assets, water and sanitation indicators, cooking fuel, dwelling quality, water collection time, household-head characteristics, and related household-context measures. Agency variables cover employment, occupation, control over earnings, decision-making, ownership, barriers, insurance, connectivity, and media exposure. Normative variables include indicators and indices derived from attitudes toward wife beating. The complete feature dictionary and country-coverage screen are reported in Appendix~\ref{app:variable_harmonization}.

A feature was retained in the pooled modeling space only when at least 60\% of observations in \emph{every} modeling country were non-missing. Features failing this cross-country coverage threshold were excluded rather than imputed solely for the purpose of creating artificial comparability. Sixteen candidate variables with especially sparse cross-country coverage were therefore removed from the pooled feature set; the complete screening table is given in Appendix~\ref{app:variable_harmonization}.

\subsection{Survey Weights and Analysis Units}

DHS weights are retained for population-oriented descriptive summaries and profile-composition analyses. The predictive models themselves operate on the modeling rows and use the explicitly defined feature and target variables rather than attempting to turn the machine-learning evaluation into a design-based estimator. This distinction is maintained throughout the paper: weighted quantities describe population composition, while predictive metrics describe held-out model performance.

The household nature of DV-module selection also affects validation. Women belonging to the same primary sampling unit can share the sampling context that determines selection. To reduce this leakage, the supervised models use a group-aware holdout based on a country--primary-sampling-unit identifier rather than a random row-wise split.

\section{Methodology}

The full analytical pipeline is documented in Appendix~\ref{app:methodology}. The main text summarizes only the components needed to understand the empirical results.

\subsection{Profile Learning}

We first construct profiles independently of the DV-selection target. The profile feature space contains 51 numeric and 5 categorical characteristics, while design variables associated directly with DV-module selection are excluded. The continuous feature representation is standardized and transformed using principal component analysis (PCA). Nineteen components are retained, capturing 81.2\% of the observed variance.

Candidate cluster counts from $K=2$ through $K=10$ are evaluated using K-means inertia, silhouette score, and Gaussian-mixture BIC. Silhouette is maximized at $K=3$, whereas GMM BIC is minimized at $K=10$. Rather than presenting either criterion as uniquely determining the final solution, the analysis uses six profiles as the reporting solution to obtain interpretable population groups while retaining the complete diagnostic comparison in Appendix~\ref{app:profile_results}.

K-means with six clusters is fitted on a 199,853-woman clustering sample and then applied to the full 796,374-woman modeling population. The resulting six profiles are:

\begin{itemize}
\item \textbf{P0}: middle rural no/low-education high-parity;
\item \textbf{P1}: richer urban higher-education;
\item \textbf{P2}: middle rural secondary;
\item \textbf{P3}: poorer rural secondary never-married;
\item \textbf{P4}: poorer rural no/low-education high-parity;
\item \textbf{P5}: richer urban higher-education never-married.
\end{itemize}

Together these profiles provide a compact representation of recurring demographic--socioeconomic configurations. Their country-specific composition and full centroids are reported in Appendix~\ref{app:profile_results}.

\subsection{Supervised Selection Models}

The supervised task predicts $y_{\mathrm{selected}}$ using a model zoo containing 22 learners spanning linear models, naive Bayes, trees, ensembles, boosting methods, kernel and distance methods, and PCA-based alternatives. XGBoost and LightGBM are included among the boosting models. The primary evaluation uses a cluster-aware train/test split based on country and primary sampling unit.

We report ROC-AUC and PR-AUC as the principal discrimination metrics, with balanced accuracy, F1, precision, recall, and Brier score retained as supplementary diagnostics. Because the selected class is common in the pooled modeling universe, PR-AUC is reported alongside ROC-AUC rather than relying on accuracy alone.

The key methodological comparison is not the absolute AUC of the strongest model, but the change in performance as feature blocks are added:
\begin{equation}
\text{Design}
\rightarrow
\text{Design + Country}
\rightarrow
\text{+ Woman}
\rightarrow
\text{+ Household}
\rightarrow
\text{+ Agency}
\rightarrow
\text{+ Norms}
\rightarrow
\text{+ Interactions}.
\end{equation}

This nested design is intended to answer whether demographic and socioeconomic information adds predictive structure after the household sampling mechanism and country context have already been supplied. The complete model zoo and all diagnostics are reported in Appendix~\ref{app:selection_models}.

\begin{table}[!ht]
\centering
\caption{Incremental predictive value of feature blocks for DV-module selection.}
\label{tab:main_nested}
\small
\begin{tabular}{lrrr}
\toprule
Feature representation & HistGradientBoosting & Logistic & LightGBM \\
\midrule
Design & 0.8287 & 0.8264 & 0.8285 \\
Design + country & 0.8516 & 0.8479 & 0.8511 \\
* woman characteristics & 0.9161 & 0.9013 & 0.9156 \\
* household characteristics & 0.9166 & 0.9020 & 0.9144 \\
* agency & 0.9167 & 0.9021 & 0.9139 \\
* norms & 0.9167 & 0.9021 & 0.9139 \\
* explicit interactions & 0.9165 & 0.9022 & 0.9144 \\
  \bottomrule
  \end{tabular}

\vspace{1mm}
\begin{minipage}{0.96\linewidth}
\footnotesize
\textit{Note.} Entries are held-out ROC-AUC from the same cluster-aware evaluation split. The main comparison is between ''Design + country'' and the subsequent feature blocks. Full PR-AUC values and all 22 learners are reported in Appendix~\ref{app:selection_models}.
\end{minipage}
\end{table}

\subsection{Interaction Analysis}

To address the interaction component of the research question, the study constructs explicit two-way product terms among education, wealth, residence, age, union status, household eligibility, agency, norms, parity, and household size. Ten terms are estimable in the final inference model. A logistic interaction model provides odds ratios with Wald confidence intervals, while the same interaction block is added to the nested predictive models.

This two-part design distinguishes statistical association from predictive contribution. A statistically nonzero interaction need not produce a large increase in out-of-sample discrimination. Accordingly, interaction terms are interpreted primarily through the combination of estimated associations, partial-dependence analyses, and held-out model comparisons; detailed results are provided in Appendix~\ref{app:interactions}.

\subsection{Cross-Country Generalization}

The pooled analysis can conceal national differences. We therefore fit the same LightGBM feature specification separately within each country and compare its ROC-AUC against the corresponding design-only baseline. We additionally perform leave-one-country-out transfer experiments in which the model is trained using the remaining four countries and evaluated on the held-out country.

This produces two complementary quantities: within-country profile gain over design, and the gain retained when the country has not been observed during training. The full country-level results and feature-ranking correlations are reported in Appendix~\ref{app:cross_country}.

\section{Results}

\subsection{Distinct Population Profiles}

The unsupervised analysis identifies six interpretable demographic--socioeconomic profiles. P2 is the largest profile, representing 23.4\% of the modeling population, followed by P0 at 20.3\%, P1 at 16.7\%, P4 at 15.6\%, P3 at 13.8\%, and P5 at 10.3\%. The profiles are differentiated by combinations of wealth, residence, educational attainment, marital status, and parity rather than by a single dimension.

The country composition of these profiles is highly uneven. P4 is the dominant profile in Afghanistan, Myanmar, and Pakistan; P2 is dominant in India; and P0 is dominant in Nepal. The weighted share of P4 ranges from approximately 83.0\% in Afghanistan to 9.9\% in Nepal, while P3 ranges from approximately 0.1\% in Afghanistan to 18.6\% in Myanmar. The complete country-by-profile matrix is provided in Appendix~\ref{app:profile_results}.

These results establish that the South Asian modeling population contains common multivariate structures but that their population prevalence is strongly country-dependent. This provides the basis for testing whether the relationship between profile characteristics and DV-module selection is similarly heterogeneous.

\subsection{Machine Learning Predicts DV-Module Selection Beyond the Design Baseline}

The principal machine-learning result is that demographic and socioeconomic information provides substantial predictive information about the realized DV-module selection outcome beyond the household design and country context.

Across 22 evaluated learners, HistGradientBoosting performs best, achieving ROC-AUC 0.9167 and PR-AUC 0.9612 on the held-out clusters. The design-only baseline achieves ROC-AUC 0.8475 and PR-AUC 0.9203. Thus, the complete feature space improves ROC-AUC by 0.0692 and PR-AUC by 0.0409. XGBoost and LightGBM are close behind, with ROC-AUC values of 0.9154 and 0.9140, respectively. The complete model comparison, including F1, precision, recall, Brier score, confusion matrices, ROC curves, and precision--recall curves, is provided in Appendix~\ref{app:selection_models}.

The nested feature-block analysis gives a more informative result than the overall AUC. With design and country included, HistGradientBoosting achieves ROC-AUC 0.8516. Adding woman-level characteristics increases this to 0.9161, while subsequent household, agency, and normative blocks produce only small additional changes. Across HistGradientBoosting, logistic regression, and LightGBM, the average gain from adding the broader demographic-socioeconomic profile space beyond design and country is 0.0608 ROC-AUC (Table~\ref{tab:main_nested}). This demonstrates that the strong predictive performance of the full model is not explained entirely by the household selection rule.

\subsection{Feature Contributions Concentrate in Design and Woman-Level Characteristics}

The feature-block attribution analysis provides a complementary view. Permuting the design block reduces ROC-AUC by 0.09972, while permuting the woman-level block reduces it by 0.09721. Country contributes a smaller but still substantial decrease of 0.04622. By comparison, the aggregate permutation effects of the interaction, household, agency, and norms blocks are much smaller: 0.00570, 0.00319, 0.00010, and effectively no additional aggregate signal, respectively.

This pattern suggests that the primary predictive structure consists of two layers. The first is the known household selection mechanism encoded by the design features. The second is the demographic and socioeconomic composition of the women to whom that design is applied. The ML results therefore support the central profile hypothesis without implying that every feature block contributes equally.

\subsection{Explicit Interactions Provide Limited Additional Out-of-Sample Gain}

Ten explicit two-way interaction products were evaluated in the interaction analysis. Seven reach the conventional 5\% significance threshold in the interaction logistic model, with the strongest estimated associations involving age-by-union status, eligible-women-by-education, eligible-women-by-wealth, wealth-by-education, urban-by-wealth, and eligible-women-by-urban residence. The detailed odds ratios and confidence intervals are reported in Appendix~\ref{app:interactions}.

Their incremental predictive contribution, however, is small. Adding the interaction products to the additive logistic model increases held-out ROC-AUC by 0.0027 and PR-AUC by 0.0008. In the nested boosting models, the final interaction block changes ROC-AUC by $-0.0002$ for HistGradientBoosting, $+0.0001$ for logistic regression, and $+0.0005$ for LightGBM.

Thus, the data support the presence of interaction structure, but explicit pairwise products are not the primary source of predictive performance. Much of the useful joint information is already accessible through the underlying demographic--socioeconomic variables and flexible nonlinear learners.

\subsection{The Profile Signal Is Strongly Country-Dependent}

Country-specific LightGBM models reveal substantial heterogeneity in the incremental value of profile information. India has the strongest within-country performance, with ROC-AUC 0.9196 compared with 0.8486 for its design baseline, giving a profile gain of +0.0710. Afghanistan is approximately aligned with its design baseline (0.8713 versus 0.8746; gain $-0.0033$), while the profile representation performs below the corresponding design baseline in Pakistan, Nepal, and Myanmar, with gains of $-0.0166$, $-0.0311$, and $-0.0344$, respectively.

The relationship among countries is not entirely unrelated: country-specific feature-importance rankings have a mean off-diagonal Spearman correlation of 0.791, with pairwise correlations ranging from 0.578 to 0.925. Nevertheless, similar feature rankings do not imply similar predictive strength. The detailed country-specific model results appear in Appendix~\ref{app:cross_country}.

Leave-one-country-out transfer produces a mean ROC-AUC transfer penalty of only $-0.0061$, indicating that the broad predictive structure transfers reasonably well in absolute terms. However, the transferred profile model does not consistently improve on the country-specific design baseline: the mean transfer gain over design is $-0.0076$. This distinction is important. The common feature structure is transferable, but the additional profile information beyond the survey design is not uniformly beneficial in unseen countries.

\subsection{Completion Is a Distinct Second-Stage Problem}

Among the women selected for the DV module, the study defines a second target corresponding to whether the interview was actually completed. Within a 300,119-woman selected subsample, 20.37\% were interviewed in the conditional modeling analysis. The best conditional completion model, LightGBM, achieves ROC-AUC 0.6638. Privacy could not be obtained for approximately 0.60\% of selected women.

These results are not part of the primary selection model because completion occurs after the sampling decision. We therefore treat completion as a separate fieldwork process rather than combining selection and interview completion into a single outcome. Detailed completion, privacy, and non-response results are reported in Appendix~\ref{app:completion}.

\section{Discussion}

This study examines whether demographic--socioeconomic characteristics provide predictive information about DHS domestic-violence (DV) module selection beyond household design and country context. Adding woman-level characteristics increases HistGradientBoosting ROC-AUC from 0.8516 to 0.9161, with an average gain of 0.0608 across the principal model families. Feature attribution further highlights the contributions of household design, woman-level characteristics, and country context. These findings indicate that the observed selection process has measurable structure beyond the supplied design baseline. However, predictive importance does not establish causality or imply differences in domestic-violence risk, as the target $V044$ represents module selection and fieldwork status rather than violence experience.

Country-specific results demonstrate substantial heterogeneity in incremental predictive value. India exhibits a positive ROC-AUC gain of 0.0710 over its design-only baseline, whereas Afghanistan shows a negligible negative gain and Pakistan, Nepal, and Myanmar show larger negative gains. Although feature-importance rankings are correlated across countries, the incremental predictive benefit does not generalize uniformly. Leave-one-country-out results similarly suggest that broad predictive patterns can transfer without consistently outperforming the design-only baseline. These findings emphasize the importance of evaluating models within national contexts rather than assuming that pooled performance represents equivalent predictive value across countries.

The six-profile solution provides an interpretable summary of demographic--socioeconomic configurations whose prevalence varies considerably across countries. However, these profiles are descriptive clusters, and the predictive gains from underlying feature blocks do not establish the independent value of profile assignments. The interaction analysis likewise shows that detectable pairwise associations do not necessarily yield substantial additional predictive performance. Flexible tree-based models can capture nonlinear relationships without extensive manual interaction engineering. Separately analyzing module selection and interview completion clarifies the distinction between sampling procedures and subsequent privacy or fieldwork constraints.

Overall, profile-based machine learning offers a complementary diagnostic framework for examining the observation process in sensitive surveys. Rather than replacing design-based inference, it helps characterize how population composition, household selection procedures, and national context relate to entry into and completion of the DV module. Distinguishing the broader women's survey population, women selected for the sensitive module, and women who complete the interview is essential for interpreting the resulting data and understanding cross-country differences in representation.

\subsection{Limitations and Future Work}

The six-profile representation provides a compact, interpretable account of demographic--socioeconomic heterogeneity while retaining sufficient detail to examine domestic-violence module selection. The choice of six profiles reflects this interpretability objective, rather than a claim that one clustering solution is universally optimal. The profiles summarize population composition, and their interpretation can be further examined across survey rounds and alternative clustering solutions. India's substantial contribution to the pooled sample reflects the scale of its national survey, not a requirement that countries contribute equally. Thus, pooled performance and feature attribution primarily characterize the combined sample, while country-specific analyses reveal variation across national settings. Future work could complement these perspectives with country-balanced and hierarchical evaluations, quantify uncertainty in country-level comparisons, and assess the reproducibility of profiles and feature attributions across surveys.

Variation in incremental predictive performance across countries is an important finding: demographic--socioeconomic characteristics provide additional predictive information beyond the design-based baseline in some settings, but not uniformly across all countries. Further research could investigate the roles of population composition, survey implementation, and country-specific selection procedures. Cross-country interpretation also requires clear distinctions among household-level subsampling, individual eligibility and selection, and subsequent interview completion. Consistent definitions of the corresponding denominators and sampling probabilities are essential for comparable rates and valid interpretation. Finally, $V044$ captures module-selection and fieldwork status, not domestic-violence experience or individual risk. The findings therefore inform understanding of selection into sensitive survey modules and the resulting observation process, rather than supporting conclusions about domestic-violence prevalence or its determinants.

\section{Conclusion}
This study uses interpretable machine learning to audit the composition and selection of the DHS domestic-violence module across South Asia. We identify six demographic--socioeconomic profiles and show that profile information provides substantial predictive value beyond the household sampling design and country, while its incremental contribution varies across countries. The results also indicate that explicit interactions add limited out-of-sample improvement, suggesting that much of the relevant structure is captured by the underlying features and nonlinear models. Importantly, these findings concern selection into the DV module rather than the experience of domestic violence itself. Overall, our analysis highlights the value of profile-based auditing for understanding differential coverage and cross-country heterogeneity in sensitive survey data and provides a framework for more careful interpretation of DHS domestic-violence samples.

\section*{Data Availability}
The datasets analyzed in this study are DHS microdata provided by the Demographic and Health Surveys (DHS) Program. The DHS Program governs access to the underlying survey files and must be obtained directly through its data-access process and applicable terms of use. Because redistribution of the underlying microdata is not permitted, we do not provide copies of the DHS datasets. To support reproducibility, we provide the analysis code, variable harmonization specifications, modeling procedures, and derived non-identifying results required to reproduce the reported analyses using appropriately authorized DHS data.

\section*{Declaration of LLM usage}
During the preparation of this manuscript, the authors used LLMs, including ChatGPT, Claude, DeepSeek, and Kimi, to assist with language refinement, paraphrasing, literature searching, and code-related support. All AI-generated outputs were carefully reviewed, verified, and edited by the authors. The authors take full responsibility for the content of this manuscript.

\bibliographystyle{plainnat}
\bibliography{references}

\newpage
\appendix


\section{Related Work}
This appendix provides the fuller literature context underlying the study, positioning the contribution relative to domestic-violence measurement, DHS-based IPV research, multivariate profiling, survey selection and non-response, interpretable machine learning, and cross-country generalization. Rather than proposing a new prediction algorithm, this study combines established analytical methods to investigate whether demographic--socioeconomic characteristics systematically structure entry into a sensitive survey module.

\subsection{Domestic-Violence Measurement in Population Surveys}

Research on violence against women requires specialized measurement procedures because disclosure may expose respondents to social, emotional, or physical risks. WHO guidance therefore emphasizes privacy, confidentiality, interviewer preparation, participant safety, and responsible dissemination of findings \citep{who2001,ellsberg2005,who2023}. These requirements influence not only research ethics but also the practical conditions under which sensitive survey information can be collected, potentially affecting data completeness and comparability across settings.

The DHS domestic-violence module operationalizes these principles through dedicated sampling and fieldwork procedures. When multiple eligible women reside in a household, the module generally selects one woman for the sensitive interview, subject to the survey-specific eligibility rules, and requires adequate privacy before the interview proceeds \citep{dhs_dv_module}. DHS documentation also distinguishes the general women's sampling weight, \texttt{V005}, from the domestic-violence weight, \texttt{D005}, which accounts for the module-specific selection and response process \citep{dhs_recode6,dhs_recode7,dhs_guide8}. These distinctions demonstrate why the observed DV-module sample should not be treated as an unconditionally representative subset of the standard women's interview sample.

This observation process motivates the present study. We treat \texttt{V044} as a module-selection and fieldwork-status indicator rather than a measure of violence itself. The principal question is consequently who enters the DV-module observation process, rather than who experiences domestic violence. Separating selection from subsequent interview completion also allows us to distinguish the initial sampling decision from privacy-related and other fieldwork constraints.

\subsection{DHS Studies of IPV and Demographic--Socioeconomic Correlates}

A substantial literature uses DHS and related population surveys to investigate intimate partner violence (IPV) and its demographic, socioeconomic, and contextual correlates. In India, educational attainment and socioeconomic circumstances have been examined as individual and contextual correlates of IPV \citep{ackerson2008}. Research in other South Asian settings has considered wealth, residence, employment, partner characteristics, fertility, household conditions, and women's decision-making \citep{ziaei2014,sapkota2024}. More broadly, multi-country research has identified recurring correlates of IPV while documenting variation across populations and settings \citep{abramsky2011,devries2013}. Recent analyses of DHS data across low- and middle-income countries further demonstrate the importance of examining socioeconomic inequalities alongside cross-country heterogeneity \citep{zhao2025}.

Despite this extensive evidence, the primary analytical objective in much of this literature is outcome-centered: demographic and socioeconomic characteristics are used to explain differences in reported violence experiences. The present study addresses a complementary methodological question. Rather than using an IPV outcome as its principal prediction target, we investigate whether the composition of women and their households is associated with selection into the specialized DV module. This distinction shifts attention from the correlates of violence to the observation process through which violence data become available for analysis.

\subsection{Multivariate Profiles and Unsupervised Learning}

Profile-based methods provide a means of representing multidimensional configurations in which demographic, socioeconomic, and household characteristics occur jointly. Principal component analysis (PCA) offers a dimensionality-reduction framework for summarizing correlated variables while retaining substantial variation in the original feature space \citep{jolliffe2002}. Clustering methods can then identify recurring configurations, although their results depend on representation, scaling, distance structure, initialization, and the selected number of clusters \citep{jain2010}.

Several criteria support the evaluation and interpretation of clustering solutions. The silhouette coefficient assesses within-cluster cohesion relative to separation from other clusters \citep{rousseeuw1987}, while Gaussian-mixture models provide a probabilistic alternative to hard cluster assignments \citep{dempster1977}. Cluster-stability research further highlights the importance of evaluating whether similar structures persist under perturbations or resampling \citep{benhur2002,hennig2007}. These methods provide complementary evidence rather than a universally decisive criterion for selecting the number of profiles.

The present study combines PCA with K-means clustering to derive interpretable demographic--socioeconomic profiles. The resulting groups are treated as empirical summaries of the observed feature space, not as formally established latent sociological or clinical classes. Their substantive labels facilitate interpretation, but they do not establish that the same underlying groups would necessarily emerge in independent survey populations. External validation and systematic stability assessment therefore remain important directions for subsequent research.

\subsection{Interactions, Selection, and non-response}

Demographic and socioeconomic characteristics may relate to survey participation through both additive associations and nonlinear combinations. Conventional regression models commonly represent associations additively unless interaction terms are specified explicitly, whereas flexible tree-based learners can capture non-linearities and interactions without requiring every combination to be engineered manually. Evaluating explicit pairwise products alongside flexible models therefore helps distinguish the presence of interaction structure from its incremental contribution to predictive performance.

Survey selection and non-response provide a complementary methodological perspective. Classical work demonstrates that nonrandom sample inclusion can affect the interpretation and estimation of relationships in observed data \citep{heckman1979}. Survey-methodological research further distinguishes unit non-response from item non-response and discusses weighting and imputation strategies for addressing missing survey information \citep{brickkalton1996}. In sensitive-survey settings, the distinction is particularly important because selection into the DV module precedes the interview, while privacy constraints and other fieldwork conditions can prevent a selected respondent from completing it \citep{who2001,who2023}. Accordingly, the present study treats selection and interview completion as separate stages rather than combining them into a single outcome.

\subsection{Interpretability and Model Explanation}

Interpretable machine learning encompasses both models whose internal relationships can be examined directly and post hoc methods that explain the behavior of more complex predictors. Research on rigorous interpretability emphasizes that explanations should be evaluated against clearly specified objectives rather than assumed to be valid merely because they are intuitive or visually accessible \citep{doshivelez2017}. Model-agnostic feature-attribution approaches, including SHAP, provide additional tools for describing the contribution of individual features to model predictions \citep{lundberg2017}.

The present analysis examines predictive contributions through nested feature-block comparisons, permutation-based importance, and partial-dependence visualizations. These analyses offer complementary views of model behavior but should not be interpreted as causal explanations. Moreover, the contribution of a feature block is not equivalent to the predictive contribution of the six cluster assignments themselves. Future work could compare profile-label representations directly with their underlying demographic--socioeconomic variables and evaluate explanation stability and faithfulness using established quantitative criteria.

\subsection{Cross-Country Generalization}

Pooled models can obscure national differences in population composition, feature distributions, survey implementation, and selection mechanisms. Evidence of heterogeneity in IPV correlates across countries reinforces the need to examine relationships within their respective contexts rather than assume that a pooled association applies uniformly \citep{abramsky2011,zhao2025}. Furthermore, validation strategies that respect dependence in the data are important because conventional random splits can underestimate predictive error when observations share underlying structures \citep{roberts2017}.

We therefore complement pooled evaluation with country-specific models and leave-one-country-out transfer experiments. Country-specific evaluation examines whether additional demographic--socioeconomic information improves discrimination within each national context, whereas leave-one-country-out evaluation assesses the extent to which learned relationships transfer to an unseen country. These analyses distinguish pooled predictive performance from cross-country transferability and make heterogeneity an explicit component of the audit.

The methodological contribution is thus not a new learning algorithm, but a framework for using established profile-learning, predictive, and explanatory methods to investigate the composition and selection process of a sensitive survey module. By distinguishing population profiles, module selection, and interview completion, the study provides a basis for evaluating differential coverage while recognizing the limitations of pooled inference and cross-country generalization.


\section{DHS Data Architecture}
\label{app:dhs_architecture}

This section documents the DHS data architecture used to construct the analytic population and to establish the correspondence between household-, household-member-, and woman-level records. The analysis distinguishes Household Recode (HR), Household Member Recode (PR), and Individual Recode (IR) files because these files represent different observational units and support different analytical tasks. HR files represent households, PR files represent listed household members, and IR files represent interviewed women aged 15--49.

\subsection{Survey and Recode Inventory}

Table~\ref{tab:appendix_archive_inventory} summarizes the supplied surveys and recode types and identifies their role in the analysis. The principal machine-learning analysis is restricted to the five countries for which the domestic-violence module selection indicator required for the pooled $V044$ analysis is available: Afghanistan, India, Myanmar, Nepal, and Pakistan.

\begin{table}[H]
\centering
\caption{DHS recode inventory and role in the analysis.}
\label{tab:appendix_archive_inventory}
\small
\begin{tabular}{lllll}
\toprule
Country & Survey & Recode & Primary unit & Analysis role \\
\midrule
Afghanistan & 2015 & HR/PR/IR & Household / woman & Included \\
Bangladesh & 2022 & HR/IR & Household / woman & IR audit; excluded from $V044$ models \\
India & 2019--21 & PR/IR & Member / woman & Included \\
Myanmar & 2015--16 & PR/IR & Member / woman & Included \\
Nepal & 2022 & PR/IR & Member / woman & Included \\
Pakistan & 2017--18 & PR/IR & Member / woman & Included \\
Sri Lanka & 1987 & IR & Woman & Historical archive; excluded from models \\
\bottomrule
\end{tabular}
\end{table}

Bangladesh is retained in the archive and suitability audit but is excluded from the $V044$ selection models because the required selection information is unavailable in the supplied data. Sri Lanka is retained as an archival survey but does not contribute to the five-country selection-modeling universe.

\subsection{Household and Population Validation}

\begin{figure}[H]
\centering
\includegraphics[width=1\linewidth]{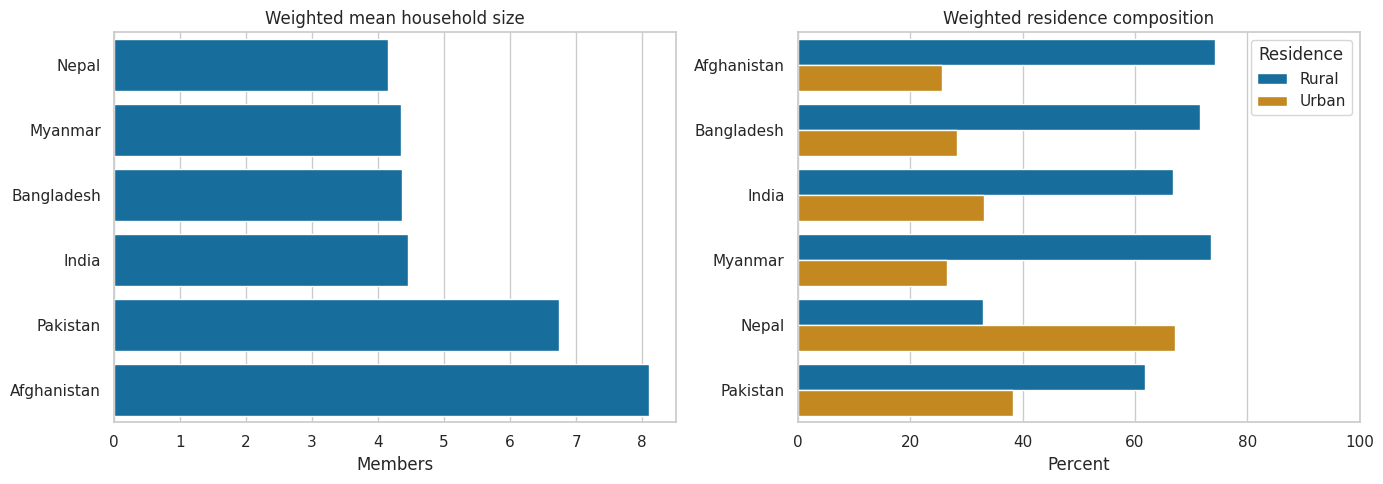}
\caption{Weighted mean household size and residence composition across countries. The left panel reports weighted mean household size, while the right panel summarizes the weighted urban--rural residence composition using the household sampling weight.}
\label{fig:app_household_size_residence}
\end{figure}

\begin{figure}[H]
\centering
\includegraphics[width=1\linewidth]{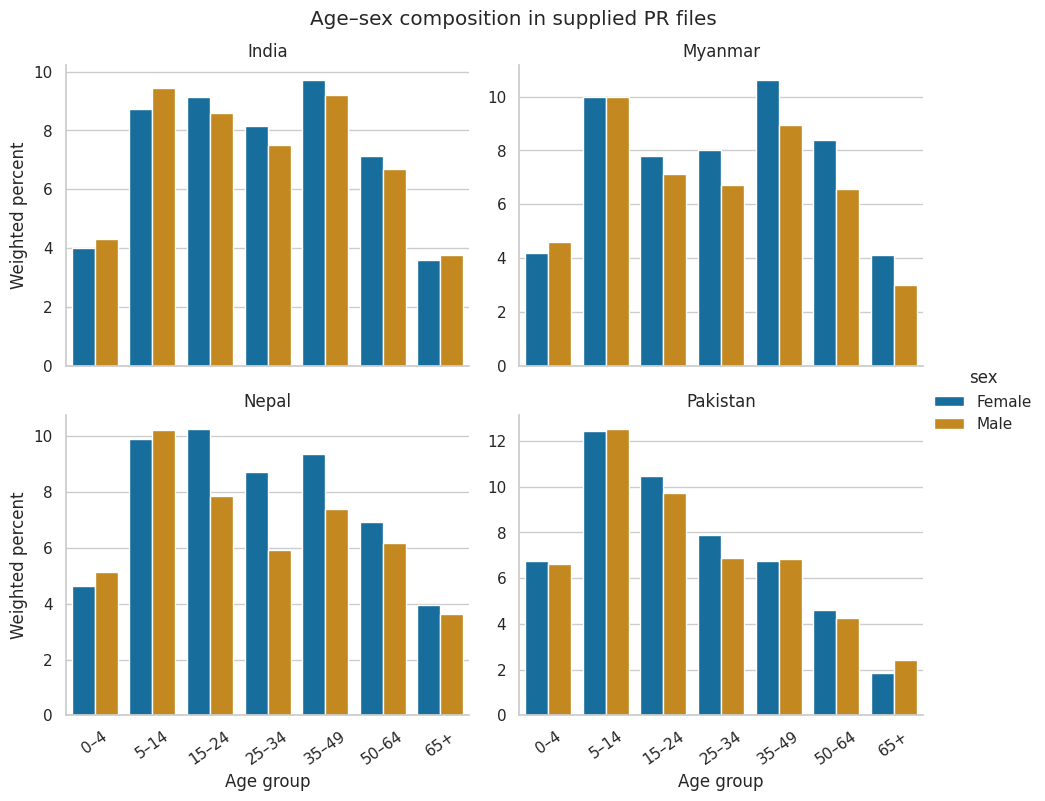}
\caption{Weighted age--sex composition of listed household members in the supplied PR files. The figure describes the household-member population represented by the available roster files and should not be interpreted as the distribution of completed women's interviews.}
\label{fig:app_age_sex_pr}
\end{figure}

\begin{figure}[H]
\centering
\includegraphics[width=1\linewidth]{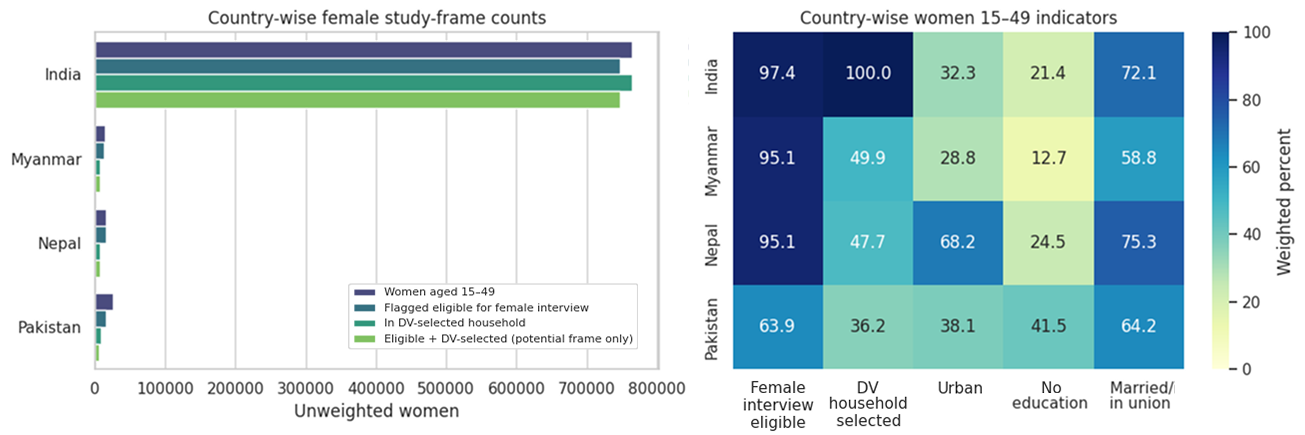}
\caption{Country-wise female study-frame size and availability of women aged 15--49 indicators. The figure documents the size of the available female study frame and the corresponding coverage of key indicators across the supplied surveys.}
\label{fig:app_country_female_frame}
\end{figure}

\begin{figure}[H]
\centering
\includegraphics[width=0.93\linewidth]{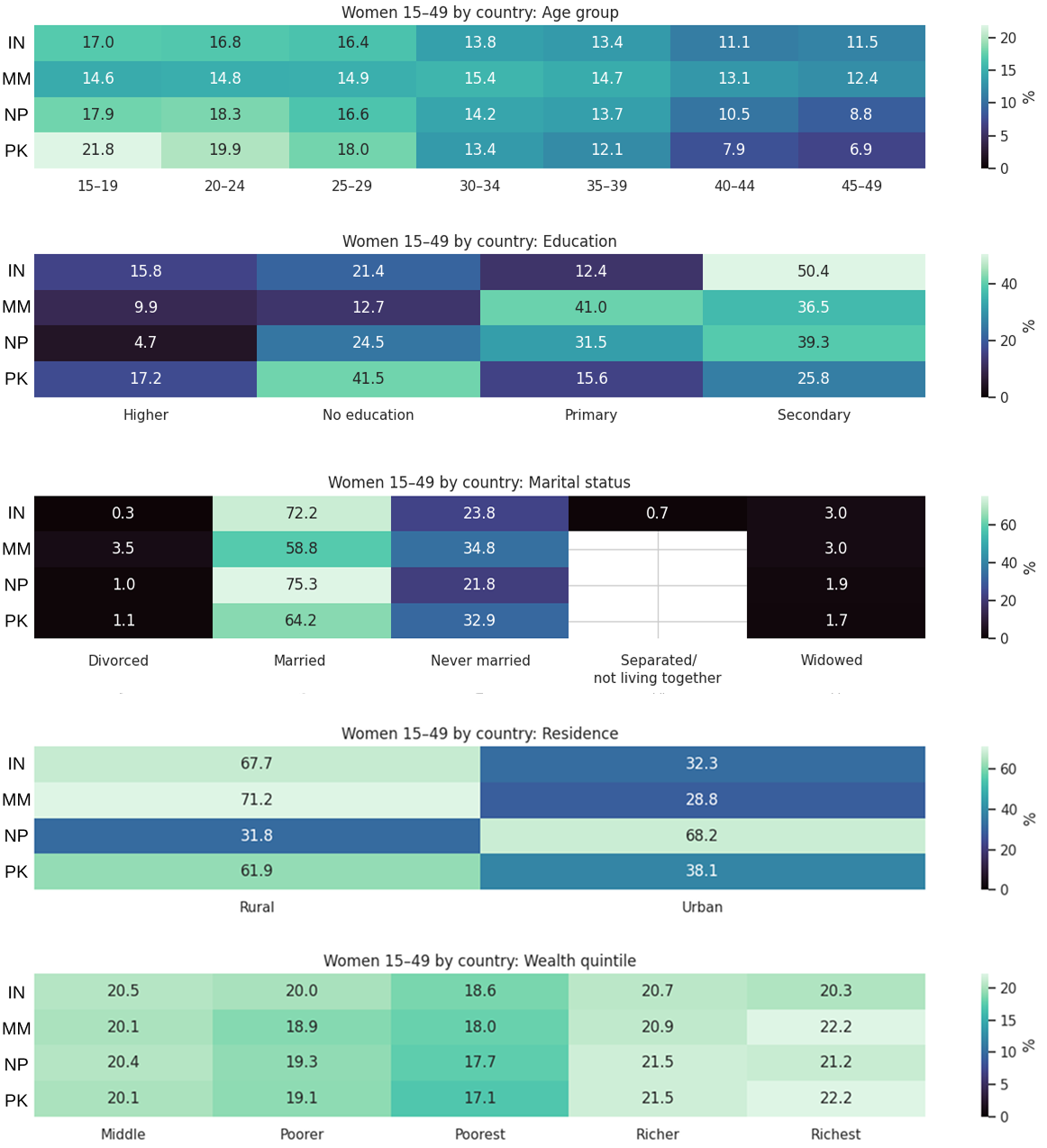}
\caption{Women aged 15--49 by country and demographic or socioeconomic dimension. Country-specific weighted distributions are shown for the dimensions used in the descriptive audit and harmonized analysis.}
\label{fig:app_women_dimension}
\end{figure}

This subsection validates the household and population samples used in the analysis. Figure~\ref{fig:app_household_size_residence} summarizes weighted household size and urban--rural composition. Figure~\ref{fig:app_age_sex_pr} describes the age--sex structure of listed household members in the PR files rather than completed individual interviews. Figure~\ref{fig:app_country_female_frame} documents the country-wise size and indicator coverage of the female study frame, while Figure~\ref{fig:app_women_dimension} presents weighted distributions across key demographic and socioeconomic dimensions. Together, these exhibits establish the structure and analytical scope of the supplied survey populations.

\subsection{Archive-Level Data Dictionary}

The study parses the DHS SAS layout files and constructs a complete data dictionary containing variable names, labels, field positions, widths, data types, and decimal specifications. The resulting inventory is used to distinguish variables that are absent from variables that are present but not asked, not applicable, or survey-specific. The dictionary provides the source-level reference for the harmonization procedures described in Appendix~\ref{app:variable_harmonization}.


\section{Variable Harmonization}
\label{app:variable_harmonization}

The pooled analysis requires comparable representations across surveys while preserving the substantive meaning of the underlying DHS variables. Harmonization is therefore performed at the variable level rather than by imposing unsupported common categories across countries. The resulting variables are organized into design, country, woman, household, agency, normative, and interaction blocks, summarized in Table~\ref{tab:appendix_feature_blocks}.

\subsection{Recode-Specific Feature Construction}

For the PR-based female frame, women are identified using $HV104=2$ and age $15$--$49$ using $HV105$. $HV117$ identifies eligibility for the women's interview, while $HV044$ identifies the household-level DV selection information in the PR-based analysis. These variables define the study frame and selection context rather than evidence of a completed interview.

For the IR-based analysis, $V005$ is the individual sampling weight, $D005$ is the domestic-violence weight, $V191$ is the wealth index, and $V044$ is used to construct the principal selection and completion targets.

\subsection{Special Codes and Scaling}

The analysis scans fixed-width DHS files in chunks and applies the implied decimal scaling specified by the DHS recode structure. Weights and the wealth index are therefore converted to their analytical scales before modeling. Survey-specific not-applicable and unknown codes are mapped to missing values rather than substantive zeros.

For the DV-frequency variables, the analysis defines an ''ever'' indicator using codes ${1,2,3,4}$ and a past-12-month indicator using ${1,2,4}$. This preserves code 4 as evidence of experience while recognizing that the corresponding frequency information is incomplete.

\subsection{Feature Blocks}

Table~\ref{tab:appendix_feature_blocks} provides the complete conceptual organization of the feature space used in the pooled selection models.

\begin{table}[H]
\centering
\caption{Feature blocks used in the pooled selection models.}
\label{tab:appendix_feature_blocks}
\small
\begin{tabular}{p{0.10\linewidth} p{0.90\linewidth}}
\toprule
Block & Principal variables and derived measures \\
\midrule
Design &
Eligible women, reciprocal eligible-women count, single-eligible-woman indicator, household size, children under five, crowding, capped eligibility categories. \\
Country &
Country and survey context, including survey year representation. \\
Woman &
Age, education, marital and union status, age at cohabitation, fertility, parity, pregnancy, child-marriage indicators, partner age and education, and related gaps. \\
Household &
Urban residence, wealth, assets, water, sanitation, fuel, dwelling quality, household-head characteristics, and relationship to household head. \\
Agency &
Employment, occupation, earnings control, household decision-making, asset ownership, healthcare barriers, insurance, mobile phone, financial accounts, internet, and media exposure. \\
Norms &
Indicators and derived indices representing attitudes toward wife beating. \\
Interactions &
Engineered two-way products involving age, union status, education, wealth, residence, empowerment, norms, household size, and eligibility. \\
\bottomrule
\end{tabular}
\end{table}

Sixteen candidate two-way products are constructed, of which ten satisfy the pooled-data coverage requirement and enter the final interaction analysis. Country-specific religion and ethnicity/caste variables are not forced into a common pooled representation when comparability cannot be established.

\begin{figure}[H]
\centering
\includegraphics[width=1\linewidth]{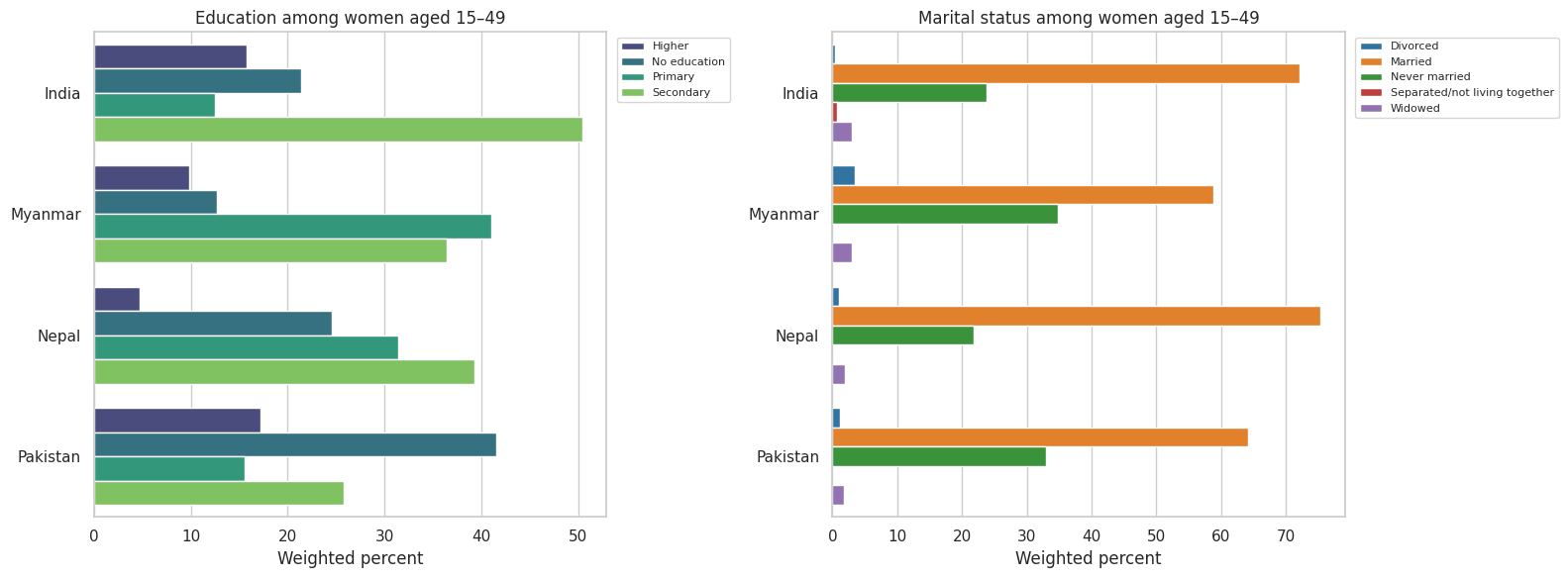}
\caption{Weighted education and marital-status composition among women aged 15--49. The panels document the cross-country distributions of two principal harmonized demographic dimensions.}
\label{fig:app_education_marital}
\end{figure}

\begin{figure}[H]
\centering
\includegraphics[width=0.90\linewidth]{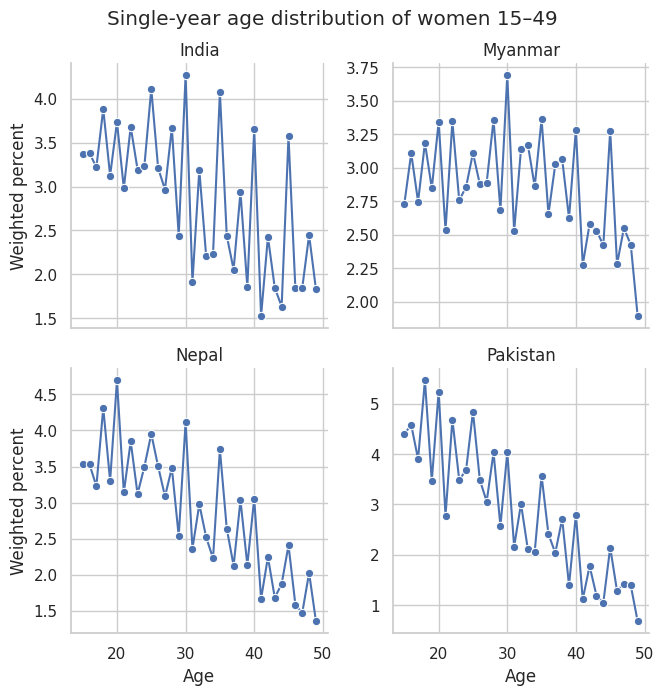}
\caption{Single-year age distribution of women aged 15--49 by country. Weighted age profiles provide a direct check on the comparability of the principal female study population across surveys.}
\label{fig:app_single_age}
\end{figure}

\subsection{Cross-Country Coverage Rule}

A candidate modeling feature is retained in the pooled matrix only when at least $60\%$ of observations in every modeling country are observed. The final pooled matrix contains 61 numeric and 8 categorical features. Variables that fail this minimum cross-country coverage requirement are removed rather than relying on extensive cross-country imputation.


\section{DV Module Analysis and Data Quality}
\label{app:dv_analysis}

This section documents the domestic-violence module as an observation and selection process rather than as a direct measure of violence exposure. It separates module selection, DV-item response coverage, and interview completion. The resulting diagnostics provide the denominator, coding, and design context required to interpret the machine-learning target.

\subsection{Selection and Interview Status}

For the primary selection analysis, the target is
\begin{equation}
y_{\mathrm{selected},i}=I!\left(V044_i \in {1,2,3}\right),
\end{equation}
where $y_{\mathrm{selected},i}=1$ indicates that woman $i$ belongs to the selected DV-module group.

For the subsequent completion analysis,
\begin{equation}
y_{\mathrm{completed},i}=I!\left(V044_i=1\right).
\end{equation}

The implemented interpretation of $V044$ is given in Table~\ref{tab:v044_status}.

\begin{table}[H]
\centering
\caption{Coding of the DHS $V044$ domestic-violence module status variable used in the analysis.}
\label{tab:v044_status}
\small
\begin{tabular}{cl}
\toprule
$V044$ & Meaning \\
\midrule
0 & Not selected for the DV module \\
1 & Selected and interviewed \\
2 & Selected but privacy was not possible \\
3 & Selected but not interviewed for another reason \\
\bottomrule
\end{tabular}
\end{table}

\subsection{Design-Based Selection Expectation}

The household-level number of eligible women, $V138$, provides the central design variable for the selection analysis. The implemented selection structure is represented as,
\begin{equation}
P(\mathrm{selected})=P(\mathrm{household\ in\ DV\ subsample})
\times
\frac{1}{V138}.
\end{equation}

The mean number of eligible women in the five-country modeling population is 1.78. Consequently, the relevant comparison is not whether demographic variables predict selection in absolute terms, but whether observed selection differs systematically from the structure already implied by household eligibility.

\begin{figure}[H]
\centering
\includegraphics[width=1\linewidth]{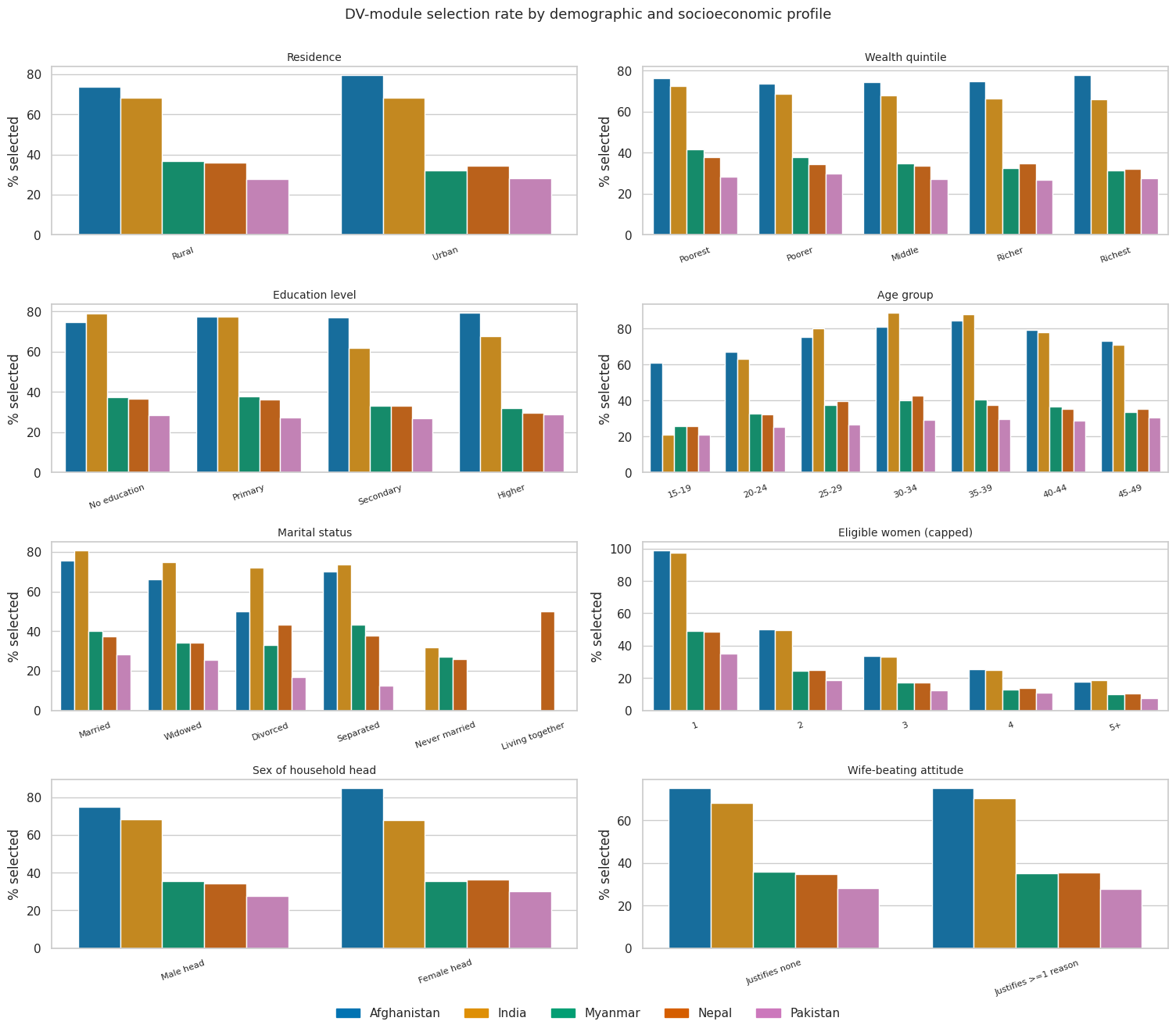}
\caption{DV-module selection rates across demographic and socioeconomic characteristics. The figure provides the descriptive selection landscape before multivariate profile learning and predictive modeling.}
\label{fig:app_selection_profile}
\end{figure}

Figure~\ref{fig:app_selection_profile} shows descriptive selection rates across demographic and socioeconomic characteristics. Figure~\ref{fig:app_v044_design} compares the observed $V044$ composition with the design expectation, while profile-specific departures from expectation are examined in Figure~\ref{fig:app_profile_selection}.

\begin{figure}[H]
\centering
\includegraphics[width=1\linewidth]{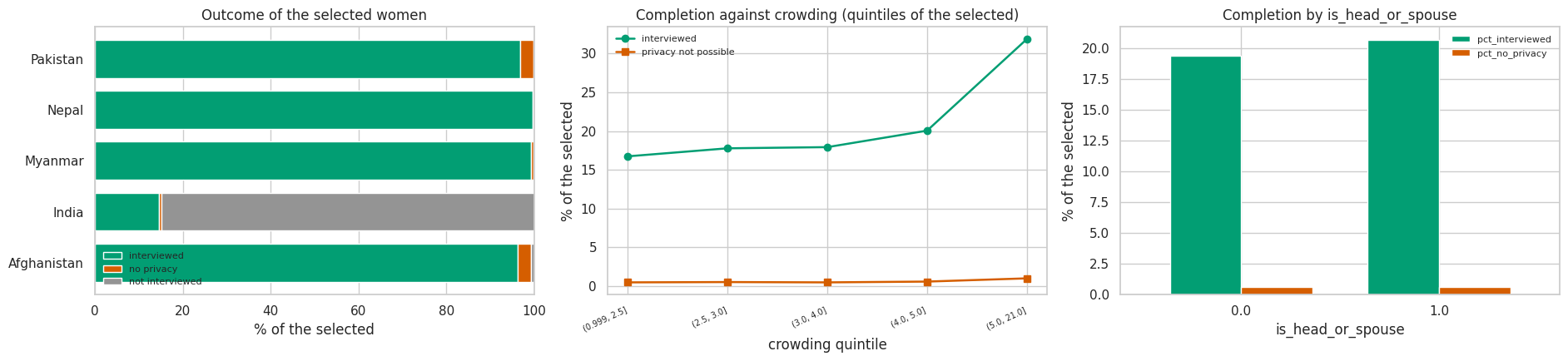}
\caption{Outcomes among women selected for the DV module and conditional completion patterns. The panels show interview completion, privacy failure, and other non-interview outcomes, together with completion patterns across household crowding quintiles and the relevant household-role characteristic.}
\label{fig:app_completion_outcomes}
\end{figure}

\subsection{Question-Level Coverage and Violence Measures}

\begin{figure}[H]
\centering
\includegraphics[width=1\linewidth]{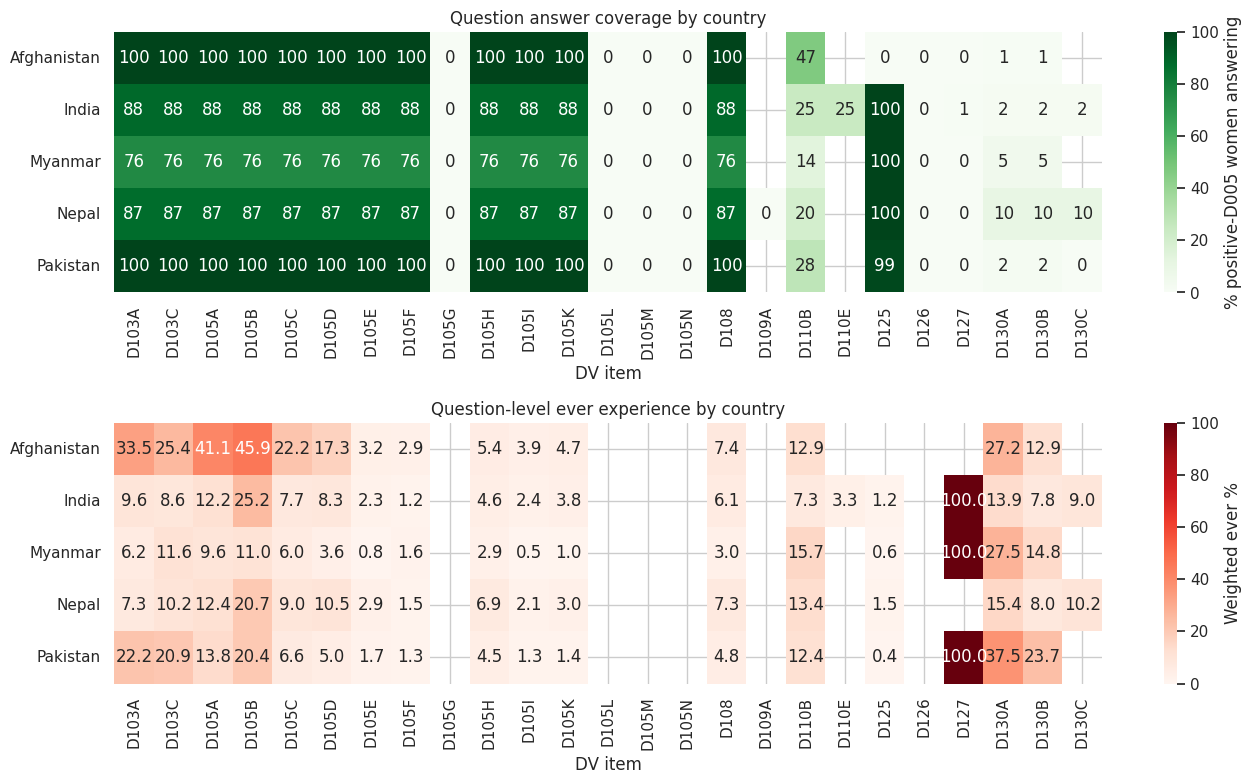}
\caption{Domestic-violence item response coverage and question-level ever-experience distributions by country. The figure separates item availability and response completeness from the primary DV-module selection analysis.}
\label{fig:app_dv_question_coverage}
\end{figure}

\begin{figure}[H]
\centering
\includegraphics[width=1\linewidth]{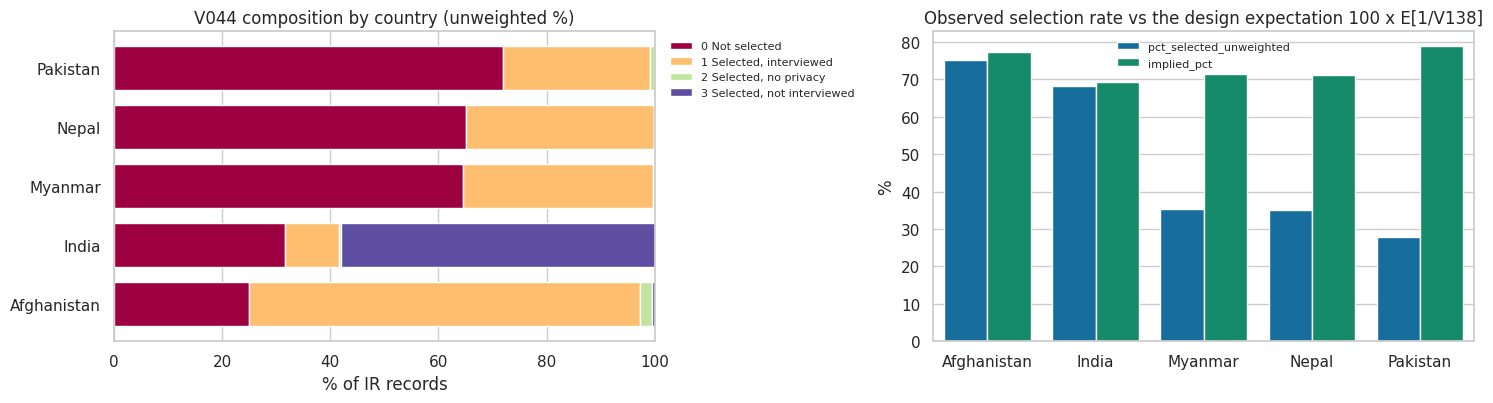}
\caption{Distribution of $V044$ and comparison of observed selection with the design expectation. The left panel reports the unweighted composition of $V044$ states by country; the right compares observed selection with $100E[1/V138]$.}
\label{fig:app_v044_design}
\end{figure}

This subsection examines DV-item availability and response coverage separately from the primary ML target. Figure~\ref{fig:app_dv_question_coverage} reports item-level response coverage and ever-experience distributions by country. Figure~\ref{fig:app_v044_design} presents the distribution of $V044$ states and the comparison between observed selection and the design expectation.

\subsection{Additional Country-Level Descriptive Tables}

Table~\ref{tab:appendix_selection_outcomes} reports the principal selection and completion outcomes for the five-country modeling universe. These values provide the country-level denominator context for the predictive analyses.

\begin{table}[H]
\centering
\caption{Principal selection and completion outcomes in the five-country modeling universe.}
\label{tab:appendix_selection_outcomes}
\small
\begin{tabular}{lrrrr}
\toprule
Country & Women & Selected & Selected (\%) & Interviewed among selected (\%) \\
\midrule
Afghanistan & 29,461 & 21,323--21,325 & 75.12\% & 96.35\% \\
India & 724,115 & 494,824--494,896 & 68.34\% & 14.6\% \\
Myanmar & 12,885 & 4,562--4,564 & 35.41\% & 99.3\% \\
Nepal & 14,845 & 5,191--5,192 & 34.97\% & 99.7\% \\
Pakistan & 15,068 & 4,216--4,217 & 27.98\% & 96.9\% \\
\midrule
Total & 796,374 & 530,933 & 66.67\% & -- \\
\bottomrule
\end{tabular}
\end{table}


\section{Completion and non-response}
\label{app:completion}

The selection analysis ends when a woman is identified as selected for the domestic-violence module. This section examines the subsequent stage of interview completion. Separating these stages allows the sampling mechanism to be distinguished from privacy and fieldwork conditions that arise after selection.

Among the 530,933 women selected for the DV module, the analysis constructs a selected-only population for the completion task. The conditional completion analysis uses 300,119 selected women, with 20.37\% classified as interviewed.

\begin{table}[H]
\centering
\caption{Prediction of interview completion under unconditional and selected-only targets.}
\label{tab:appendix_completion_models}
\small
\begin{tabular}{llrr}
\toprule
Target & Model & ROC-AUC & PR-AUC \\
\midrule
Completed, all women & Logistic regression & 0.7350 & 0.4340 \\
Completed, all women & Random forest & 0.7377 & 0.4367 \\
Completed, all women & LightGBM & 0.7332 & 0.4317 \\
\midrule
Interviewed $\mid$ selected & Logistic regression & 0.6607 & 0.5307 \\
Interviewed $\mid$ selected & Random forest & 0.6600 & 0.5272 \\
Interviewed $\mid$ selected & LightGBM & 0.6638 & 0.5330 \\
\bottomrule
\end{tabular}
\end{table}

Table~\ref{tab:appendix_completion_models} shows that conditional completion is substantially harder to predict than the unconditional completion target, which partly reflects the fact that unconditional completion retains information about the initial selection mechanism.

The analysis reports that approximately 0.60\% of selected women could not be interviewed because privacy could not be obtained. Figure~\ref{fig:app_completion_outcomes} provides the corresponding descriptive breakdown and examines completion across household crowding and household-role measures.


\section{Detailed Methodology}
\label{app:methodology}

This appendix provides the complete computational specification underlying the main-paper methodology. It documents the modeling universe, preprocessing, profile-space construction, PCA reduction, cluster selection, profile assignment, supervised-learning design, grouped validation, interaction construction, and cross-country evaluation.

\subsection{Pooled modeling Universe}

The pooled modeling dataset contains 796,374 women aged 15--49 across Afghanistan, India, Myanmar, Nepal, and Pakistan. For computationally intensive supervised-learning experiments, the analysis limits the modeling sample to approximately 300,000 observations. The resulting pooled modeling dataset contains 300,100 women and 12,280 country--PSU clusters.

A 25\% held-out test partition is constructed using cluster-aware grouping. The training set contains 225,132 women and 9,210 clusters, while the held-out test set contains 74,968 women and 3,070 clusters. No primary sampling unit occurs in both partitions. The positive-class prevalence is approximately 66.70\% in both partitions.

This grouped validation scheme is designed to reduce information leakage caused by shared survey and sampling context among women belonging to the same PSU.

\subsection{Profile Feature Space}

Profile construction deliberately excludes the direct DV-selection design the variables. The profile matrix contains 51 numeric and 5 categorical variables. After preprocessing, the transformed representation contains 79 columns.

The standardized profile representation is reduced through principal component analysis:

\begin{equation}
\mathbf{Z}=\mathbf{X}_{\mathrm{profile}}\mathbf{W},
\end{equation}

where $\mathbf{W}$ contains the PCA loading vectors and $\mathbf{Z}$ denotes the reduced feature representation.

\begin{figure}[H]
\centering
\includegraphics[width=1\linewidth]{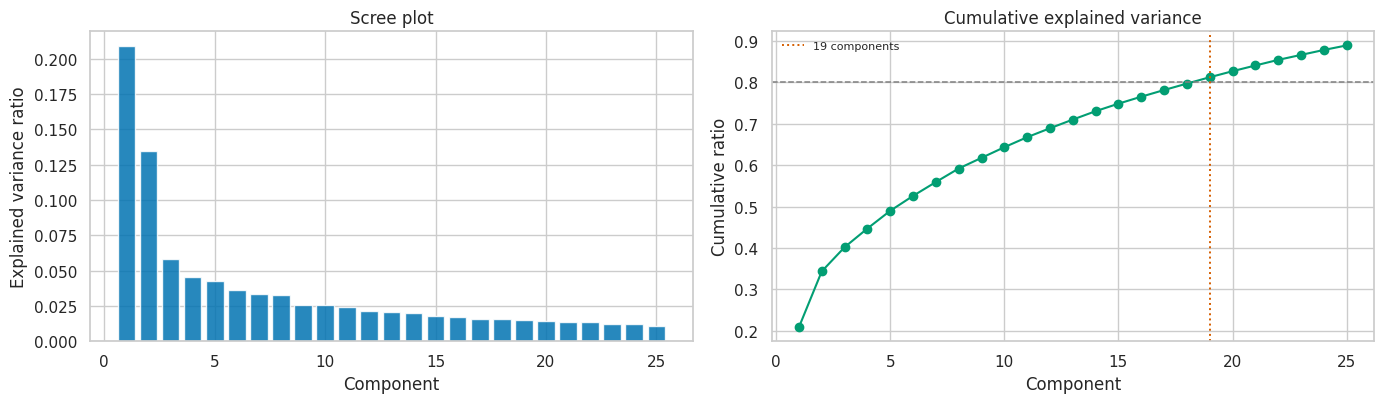}
\caption{PCA scree plot and cumulative explained variance. Nineteen principal components are retained because they explain at least 80\% of the variance in the standardized profile space.}
\label{fig:app_pca_variance}
\end{figure}

The 19 retained components explain 81.2\% of the observed variance. The full variance diagnostics are shown in Figure~\ref{fig:app_pca_variance}. The feature blocks entering the profile representation are summarized in Table~\ref{tab:appendix_feature_blocks}.

\subsection{Cluster Selection}

Candidate cluster counts $K=2,\ldots,10$ are evaluated using K-means inertia, silhouette score, Gaussian-mixture BIC, and Gaussian-mixture AIC. These diagnostics are treated as complementary evidence rather than as a single automatic criterion.

Silhouette favors $K=3$, whereas Gaussian-mixture BIC continues to improve at larger values of $K$. The analysis therefore retains $K=6$ as the reporting solution to balance empirical diagnostics with substantive interpretability. Six-cluster K-means is fitted using 199,853 women, and the resulting profile assignment is propagated to the complete 796,374-woman modeling population.

\begin{figure}[H]
\centering
\includegraphics[width=1\linewidth]{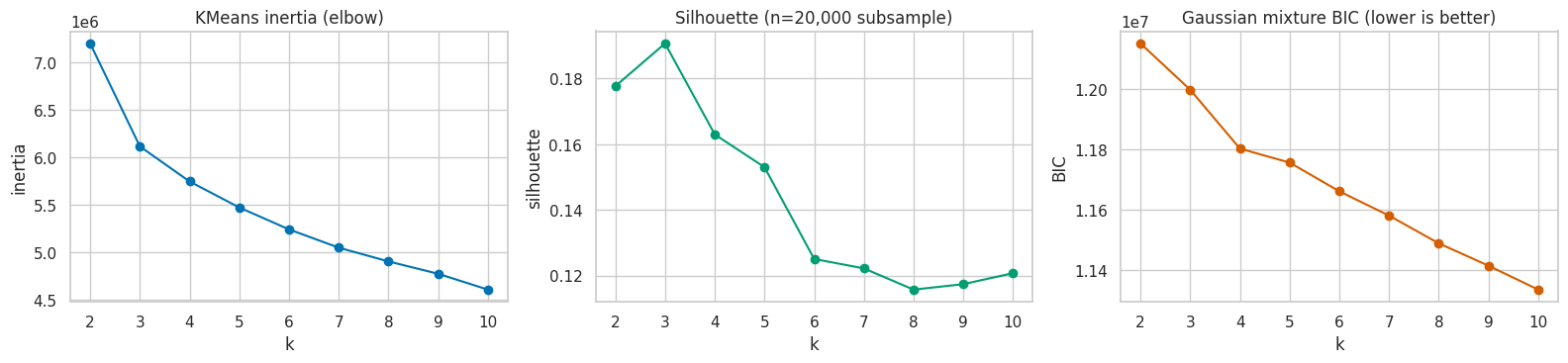}
\caption{Cluster-number diagnostics for the PCA-reduced profile representation. K-means inertia, silhouette score, and Gaussian-mixture BIC are shown across candidate values of $K$.}
\label{fig:app_cluster_diagnostics}
\end{figure}

Figure~\ref{fig:app_cluster_diagnostics} presents the complete cluster-number diagnostics.

\subsection{Profile Assignment}

The six reported profiles are interpreted from their original feature distributions rather than from PCA coordinates alone. Table~\ref{tab:profile_interpretation} summarizes the labels used throughout the main paper and appendix.

\begin{table}[!ht]
\centering
\caption{Interpretation of the six reported demographic--socioeconomic profiles.}
\label{tab:profile_interpretation}
\small
\begin{tabular}{cp{0.47\linewidth}p{0.36\linewidth}}
\toprule
Profile & Interpretation & Key cross-country pattern \\
\midrule
P0 & Middle rural, no/low education, high parity
& Most prevalent in Nepal; also substantial in Pakistan and India. \\
P1 & Richer urban, higher education
& Common in India, Nepal, and Pakistan; uncommon in Afghanistan. \\
P2 & Middle rural, secondary education
& Broadly distributed and particularly prevalent in India and Nepal. \\
P3 & Poorer rural, secondary education, never-married
& Especially represented in Myanmar and India; nearly absent in Afghanistan and Pakistan. \\
P4 & Poorer rural, no/low education, high parity
& Dominant in Afghanistan and strongly represented in Pakistan and Myanmar. \\
P5 & Richer urban, higher education, never-married
& More represented in Myanmar, Nepal, and India; nearly absent in Afghanistan and Pakistan. \\
\bottomrule
\end{tabular}
\end{table}

The profile labels are descriptive summaries of empirical clusters and are not treated as latent sociological or clinical classes.

\subsection{Country Composition}
\begin{figure}[H]
\centering
\includegraphics[width=1\linewidth]{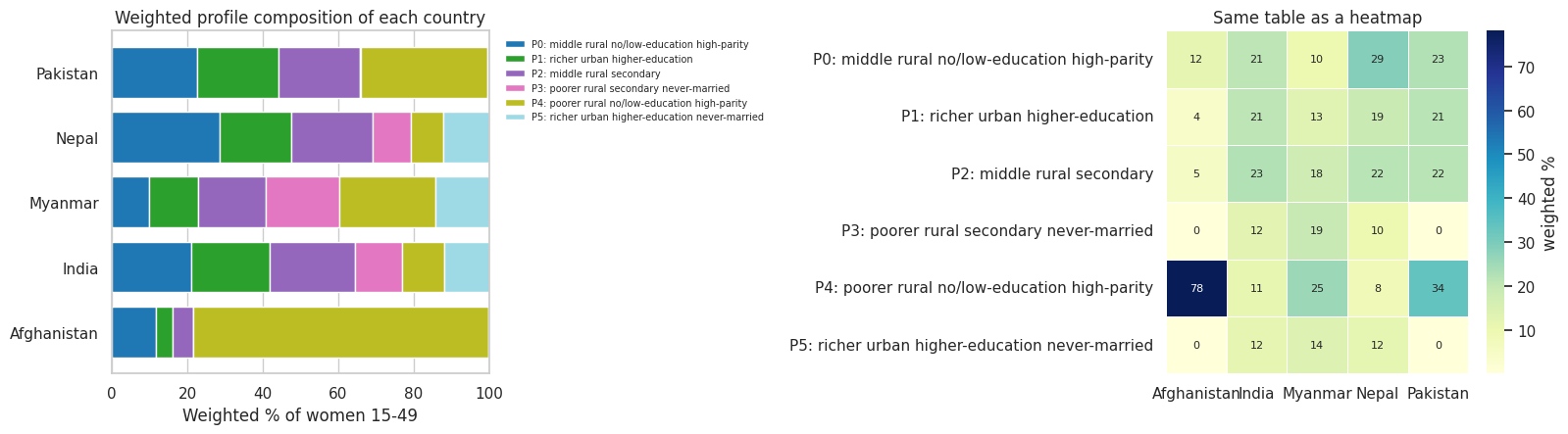}
\caption{Weighted composition of the six demographic--socioeconomic profiles within each country. The left panel reports the country-by-profile composition and the right panel presents the same values as a heatmap.}
\label{fig:app_profile_composition}
\end{figure}

\begin{figure}[H]
\centering
\includegraphics[width=1\linewidth]{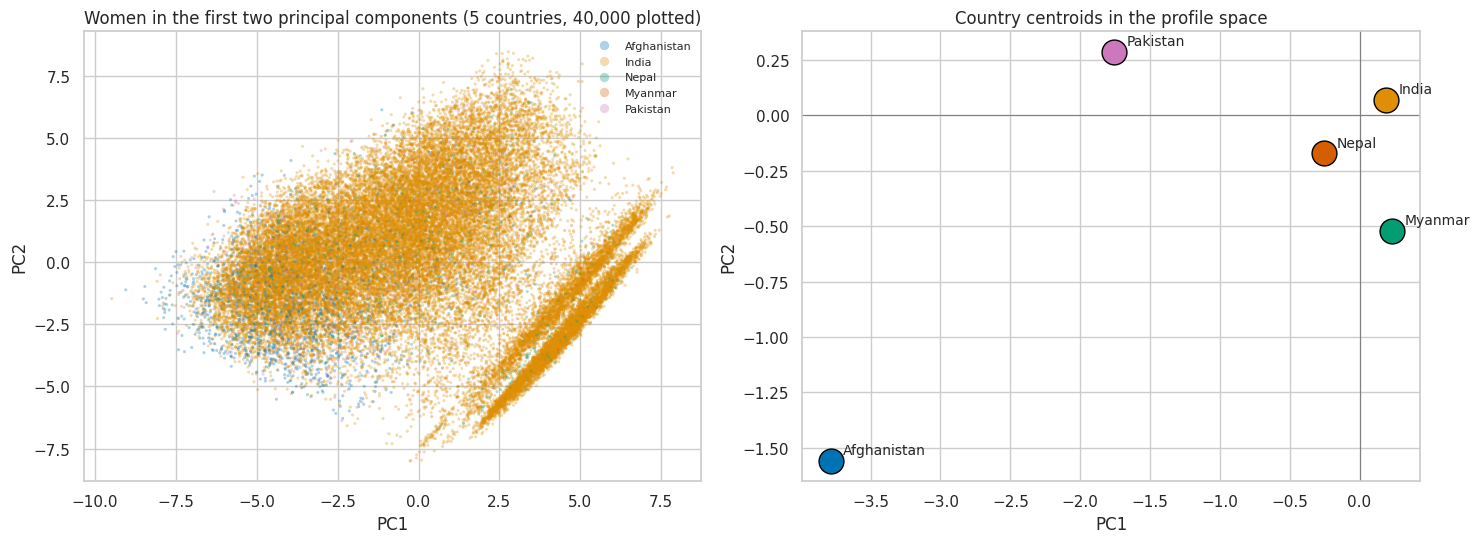}
\caption{Women in the first two principal components and country centroids in the profile space. The figure visualizes the low-dimensional representation used for profile construction and overlays the centroid of each modeling country.}
\label{fig:app_pca_country_space}
\end{figure}

\begin{figure}[H]
\centering
\includegraphics[width=1\linewidth]{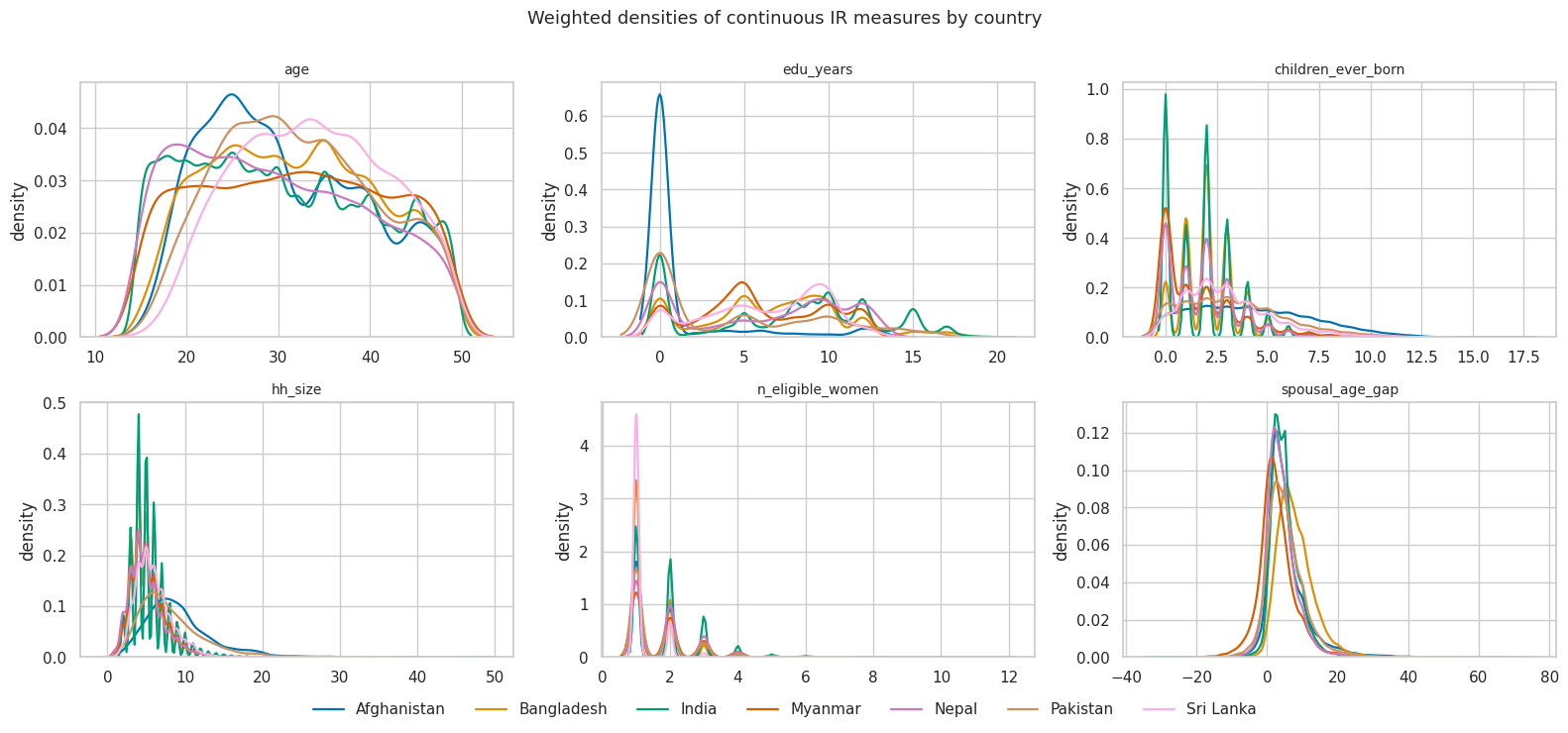}
\caption{Weighted distributions of continuous Individual Recode measures by country. Density curves provide additional descriptive context for the demographic and socioeconomic heterogeneity underlying the profile solution.}
\label{fig:app_ir_densities}
\end{figure}

This subsection examines the distribution of the six profiles across the five modeling countries. Figure~\ref{fig:app_profile_composition} presents weighted country-by-profile composition and the corresponding heatmap. Figure~\ref{fig:app_pca_country_space} shows the position of women and country centroids in the PCA space, while Figure~\ref{fig:app_ir_densities} provides weighted densities for continuous IR measures.

\subsection{Profile Selection Rates}

The profile-level selection analysis compares the observed selection rate for profile $p$,

\begin{equation}
\widehat{P}_{p}
=\frac{1}{n_p}
\sum_{i:P_i=p}
y_{\mathrm{selected},i},
\end{equation}

with the corresponding design expectation,

\begin{equation}
100\times E[1/V138\mid P_i=p].
\end{equation}

\begin{figure}[H]
\centering
\includegraphics[width=1\linewidth]{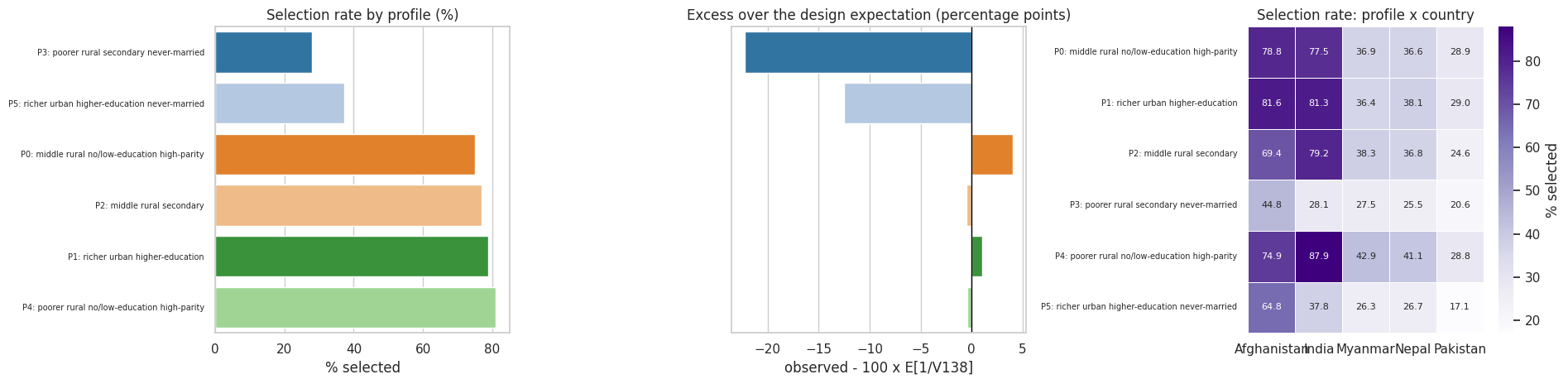}
\caption{Profile-level DV-module selection and cross-country heterogeneity. The panels show pooled selection rates by profile, deviations from the design expectation in percentage points, and the profile-by-country selection-rate matrix.}
\label{fig:app_profile_selection}
\end{figure}

The difference is reported in percentage points. Positive values indicate selection above the design expectation, while negative values indicate selection below it. Figure~\ref{fig:app_profile_selection} presents the pooled profile selection rates, deviations from expectation, and profile-by-country selection matrix.

\section{Additional Profile Results}
\label{app:profile_results}

This section reports supplementary findings concerning profile prevalence, country composition, and profile-specific selection. These results support interpretation of the principal ML findings in the main paper while preserving detailed country-level evidence in the appendix.

\subsection{Profile Composition Across Countries}

Table~\ref{tab:appendix_profile_country_mix} reports weighted country-level profile composition. The values demonstrate that the same profile can represent very different shares of the population across countries.

\begin{table}[H]
\centering
\caption{Weighted profile composition by country (\%).}
\label{tab:appendix_profile_country_mix}
\small
\begin{tabular}{lrrrrrr}
\toprule
Country & P0 & P1 & P2 & P3 & P4 & P5 \\
\midrule
Afghanistan & 11.7\% & 4.3\% & 5.4\% & 0.1\% & 78.1\% & 0.3\% \\
India & 21.0\% & 20.9\% & 22.5\% & 12.5\% & 11.4\% & 11.8\% \\
Myanmar & 9.9\% & 12.9\% & 17.9\% & 19.5\% & 25.4\% & 14.3\% \\
Nepal & 28.7\% & 19.0\% & 21.6\% & 10.1\% & 8.5\% & 12.2\% \\
Pakistan & 22.6\% & 21.5\% & 21.5\% & 0.3\% & 33.7\% & 0.4\% \\
\bottomrule
\end{tabular}
\end{table}

The full weighted profile-composition visualization is provided in Figure~\ref{fig:app_profile_composition}.

\subsection{Profile-Specific Selection}

The profile-specific selection rates and their deviations from the design expectation are reported in Figure~\ref{fig:app_profile_selection}. These results should be interpreted jointly with the $V138$ design structure rather than as unconditional demographic determinants of module selection.

\subsection{Additional Descriptive Measures}

\begin{figure}[H]
\centering
\includegraphics[width=1\linewidth]{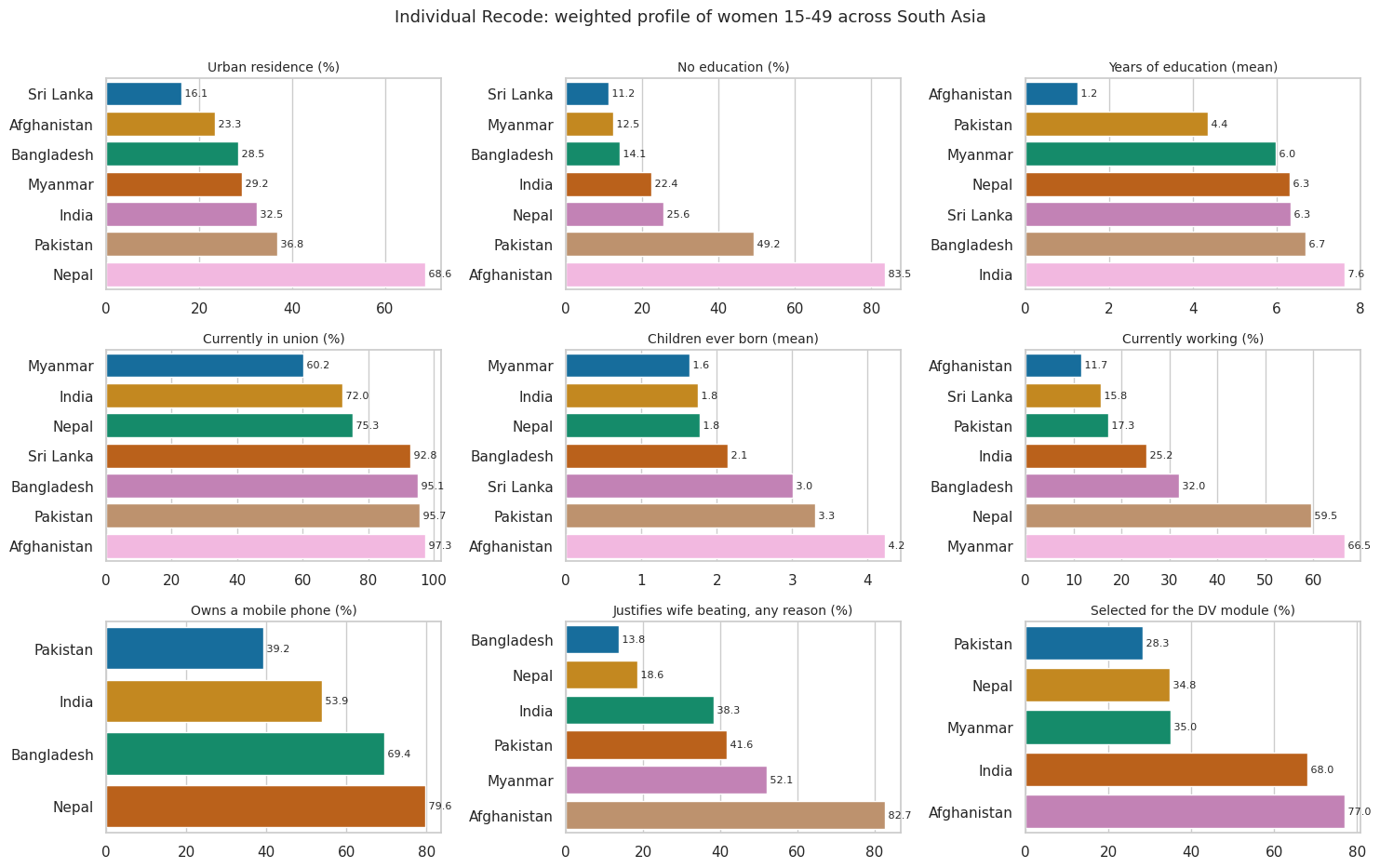}
\caption{Weighted Individual Recode profile of women aged 15--49 across the South Asian modeling countries. The figure summarizes the principal demographic and socioeconomic characteristics entering the profile analysis.}
\label{fig:app_ir_weighted_profile}
\end{figure}

\begin{figure}[H]
\centering
\includegraphics[width=1\linewidth]{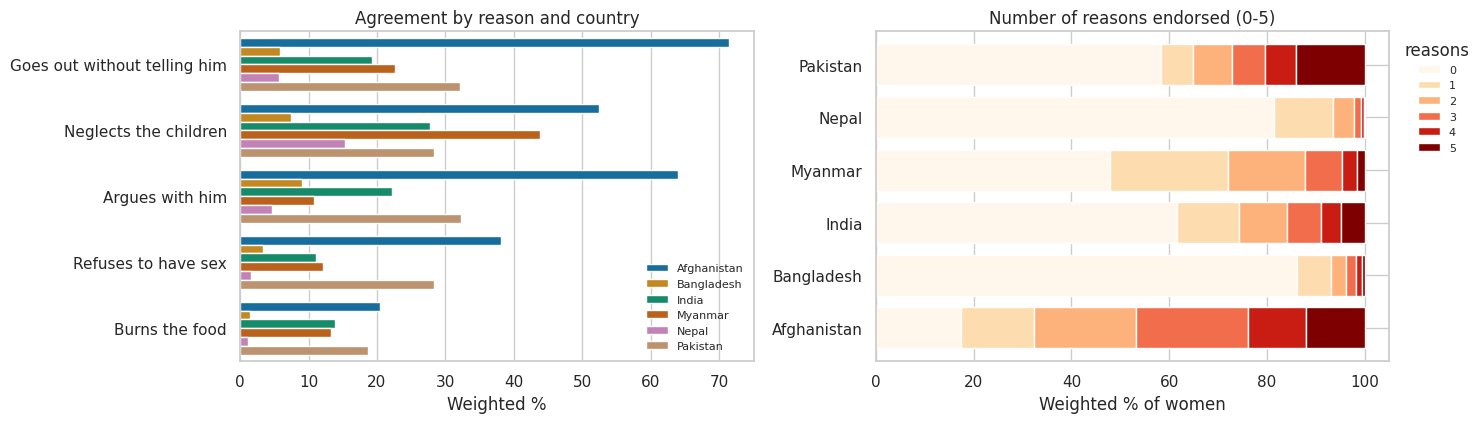}
\caption{Agreement with reason-specific indicators and number of reasons endorsed by country. The figure summarizes country-level variation in the normative indicators included in the broader feature representation.}
\label{fig:app_reason_agreement}
\end{figure}

Figure~\ref{fig:app_ir_weighted_profile} provides the weighted overall Individual Recode profile of women aged 15--49, while Figure~\ref{fig:app_reason_agreement} summarizes the country distribution of normative indicators used in the broader feature representation.

\section{Selection Models}
\label{app:selection_models}

This section presents the complete supervised-learning evidence for predicting DV-module selection. 

\subsection{Model Zoo}
Table~\ref{tab:appendix_model_zoo} reports the complete pooled learner comparison for DV-module selection. All 22 learners are trained and evaluated on the same PSU-aware training and test partition, allowing ROC-AUC, PR-AUC, and F1 to be compared directly. The learner set spans naive baselines, linear and probabilistic classifiers, tree-based ensembles, kernel methods, and PCA-based pipelines, with the strongest pooled results highlighted.

\begin{table*}[!ht]
\centering
\caption{Complete pooled learner comparison for DV-module selection.}
\label{tab:appendix_model_zoo}
\small
\begin{tabular}{lrrr}
\toprule
Model & ROC-AUC & PR-AUC & F1 \\
\midrule
Majority class & 0.5000 & 0.6670 & 0.800 \\
Stratified random & 0.5016 & 0.6677 & 0.668 \\
Design-only baseline & 0.8475 & 0.9203 & 0.784 \\
Logistic regression & 0.9022 & 0.9547 & 0.840 \\
Linear SVM & 0.9024 & 0.9547 & 0.835 \\
Linear discriminant analysis & 0.9031 & 0.9550 & 0.860 \\
Gaussian naive Bayes & 0.8535 & 0.9116 & 0.851 \\
Bernoulli naive Bayes & 0.8649 & 0.9317 & 0.832 \\
Decision tree (depth 5) & 0.9070 & 0.9461 & 0.816 \\
Decision tree (tuned) & 0.9126 & 0.9578 & 0.837 \\
Random forest & 0.9146 & 0.9602 & 0.837 \\
Extra trees & 0.9124 & 0.9591 & 0.832 \\
AdaBoost & 0.9144 & 0.9601 & 0.861 \\
Gradient boosting & 0.9159 & 0.9608 & 0.862 \\
HistGradientBoosting & \textbf{0.9167} & \textbf{0.9612} & 0.861 \\
K-nearest neighbours & 0.8865 & 0.9436 & 0.854 \\
RBF SVM & 0.8914 & 0.9511 & 0.834 \\
PCA $\rightarrow$ logistic & 0.8740 & 0.9339 & 0.836 \\
PCA $\rightarrow$ linear SVM & 0.8737 & 0.9334 & 0.834 \\
PCA $\rightarrow$ random forest & 0.8904 & 0.9462 & 0.831 \\
XGBoost & 0.9154 & 0.9608 & 0.840 \\
LightGBM & 0.9140 & 0.9597 & 0.840 \\
\bottomrule
\end{tabular}
\end{table*}

\subsection{Nested Feature-Block Analysis}

\begin{figure}[H]
\centering
\includegraphics[width=0.75\columnwidth]{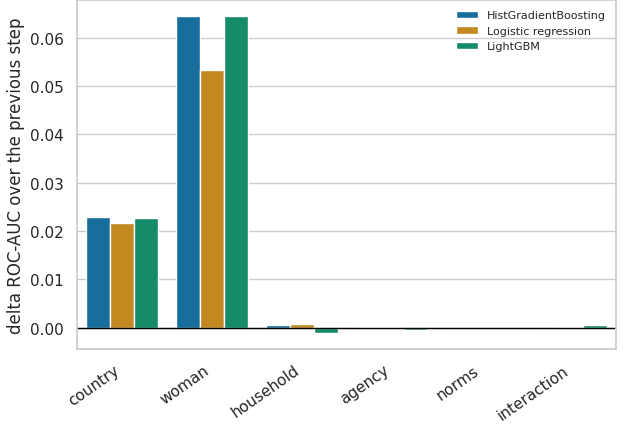}
\caption{Marginal ROC-AUC contribution of successive feature blocks. The figure shows the incremental discrimination associated with each additional feature block in the nested model ladder.}
\label{fig:marginal_roc_auc}
\end{figure}

The nested feature-block experiment evaluates the incremental information contributed by country, woman-level, household, agency, normative, and explicit interaction variables. Table~\ref{tab:predictive_evidence_profiles} reports the held-out performance at each stage. Figure~\ref{fig:marginal_roc_auc} visualizes the corresponding marginal ROC-AUC contributions.

\begin{table*}[!ht]
\centering
\caption{Incremental predictive evidence from demographic, socioeconomic, agency, normative, and interaction feature blocks. Models are evaluated on the same held-out PSUs; $\Delta$ values are relative to the design-only baseline.}
\label{tab:predictive_evidence_profiles}
\scriptsize
\begin{tabular}{lrrrrrr}
\toprule
Specification & Added block & ROC-AUC & PR-AUC & $\Delta$ROC-AUC & $\Delta$PR-AUC & Interpretation \\
\midrule
Design-only & Design & 0.8287 & 0.8986 & -- & -- & Sampling mechanism \\
Design + country & Country & 0.8516 & 0.9250 & +0.0229 & +0.0264 & Country variation \\
Design + demographics & Woman & 0.9161 & 0.9609 & +0.0874 & +0.0623 & Large additional signal \\

* household & Household & 0.9166 & 0.9611 & +0.0879 & +0.0625 & Marginal additional signal \\
* agency + norms & Agency + norms & 0.9167 & 0.9612 & +0.0880 & +0.0626 & Performance plateau \\
* interactions & Explicit interactions & 0.9165 & 0.9611 & +0.0878 & +0.0625 & Negligible incremental gain \\
  \bottomrule
  \end{tabular}
  \end{table*}

Relative to the design-plus-country specification, the addition of the broader demographic--socioeconomic representation produces the largest gain. Across HistGradientBoosting, logistic regression, and LightGBM, the mean ROC-AUC improvement from the design-plus-country specification to the full representation is 0.0608.

\begin{figure}[H]
\centering
\includegraphics[width=0.75\columnwidth]{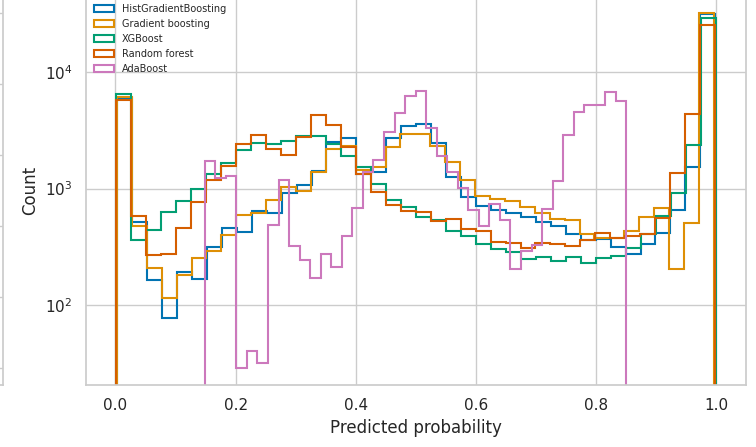}
\caption{Distribution of predicted selection probabilities across evaluated models. The figure provides an additional view of score separation beyond aggregate AUC statistics.}
\label{fig:predicted_probability_distribution}
\end{figure}

\subsection{Classification Diagnostics}

\begin{figure}[H]
\centering
\includegraphics[width=\columnwidth]{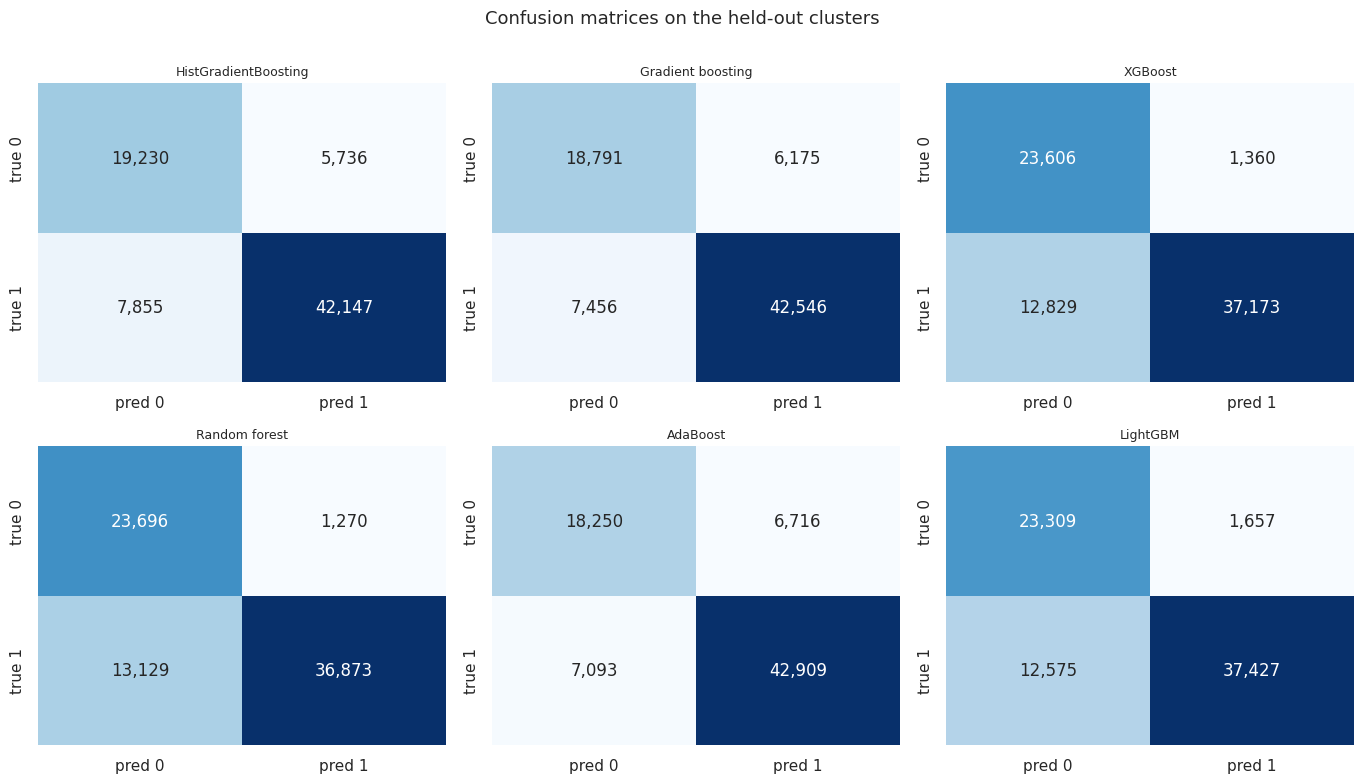}
\caption{Confusion matrices for selected high-performing DV-module selection models on held-out PSUs.}
\label{fig:app_confusion_matrices}
\end{figure}

\begin{figure}[H]
\centering
\includegraphics[width=\columnwidth]{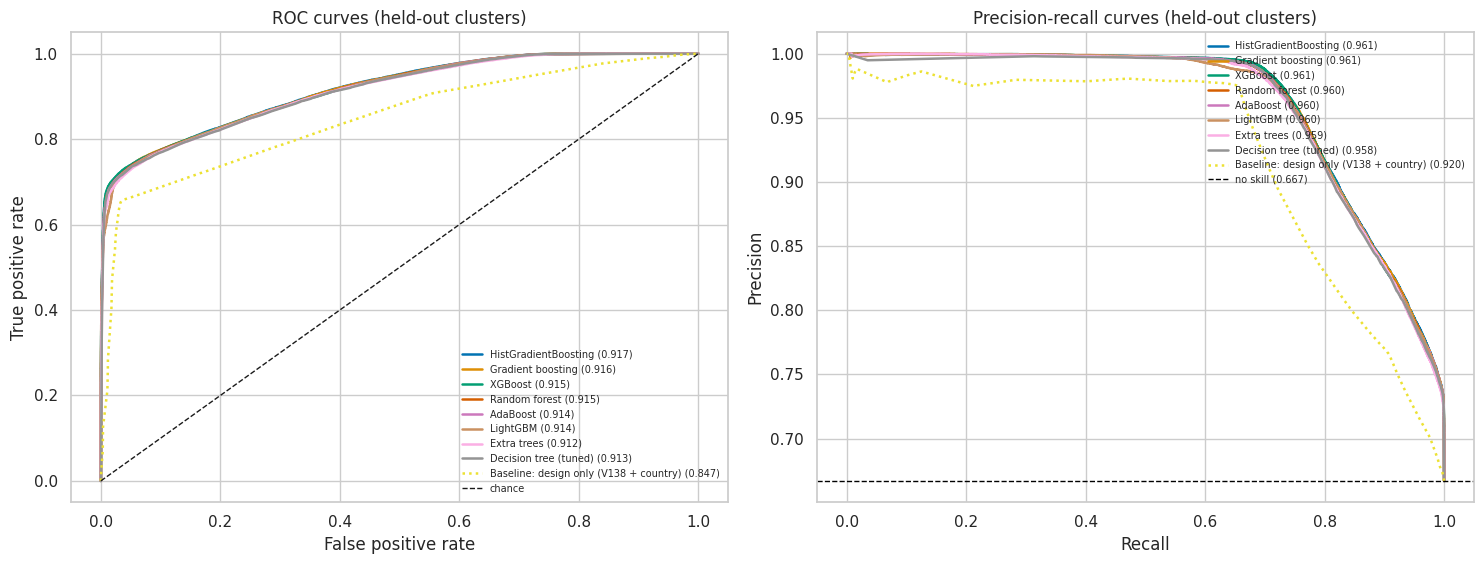}
\caption{ROC and precision--recall curves on held-out PSUs. Curves complement the learner-level AUC comparison by showing the ranking behavior of selected models across classification thresholds.}
\label{fig:app_roc_pr_curves}
\end{figure}

\begin{figure}[H]
\centering
\includegraphics[width=\columnwidth]{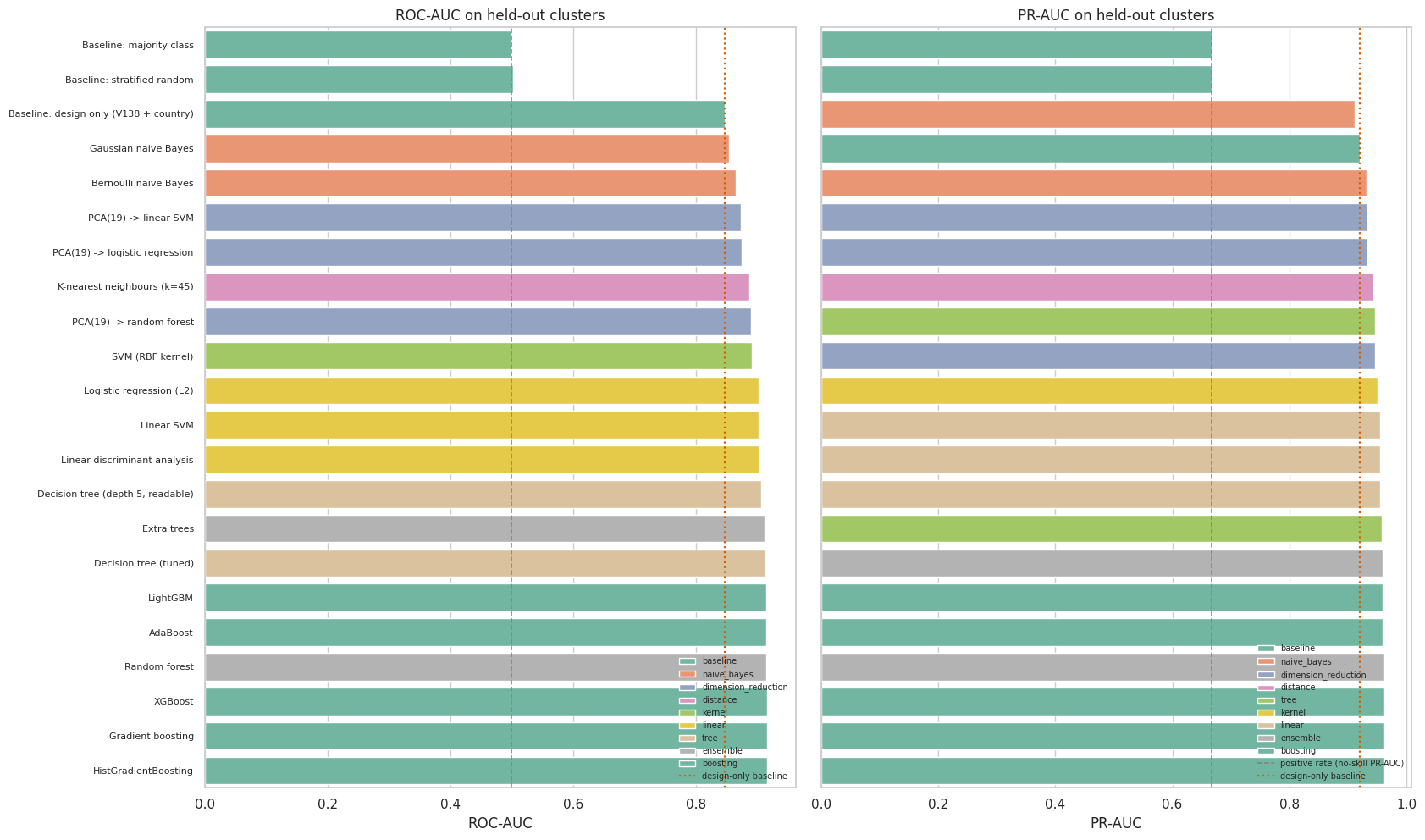}
\caption{ROC-AUC and precision--recall AUC across the complete learner set. HistGradientBoosting provides the strongest pooled discrimination in the study.}
\label{fig:app_auc_comparison}
\end{figure}

This subsection reports detailed held-out classification diagnostics. Figure~\ref{fig:app_confusion_matrices} shows confusion matrices for selected high-performing learners. Figure~\ref{fig:app_roc_pr_curves} presents ROC and precision--recall curves, while Figure~\ref{fig:app_auc_comparison} compares ROC-AUC and PR-AUC across the full learner set.

\section{Interactions}
\label{app:interactions}

This section examines whether demographic, socioeconomic, household, and design variables exhibit non-additive relationships with DV-module selection. The analysis distinguishes statistical association from predictive contribution by combining explicit product terms, logistic interaction estimates, feature-importance analysis, and held-out model comparisons.

\subsection{Engineered Interaction Terms}
Sixteen candidate products are constructed, of which ten satisfy the pooled coverage requirement and enter the final interaction analysis. Table~\ref{tab:appendix_interactions} reports the largest estimated associations. Odds ratios are expressed per standard deviation of the corresponding product term.

\begin{table}[H]
\centering
\caption{Largest estimated interaction associations with DV-module selection. Odds ratios are expressed per standard deviation of the corresponding product term.}
\label{tab:appendix_interactions}
\small
\begin{tabular}{lrrr}
\toprule
Interaction & OR & 95\% CI lower & 95\% CI upper \\
\midrule
Age $\times$ union status & 0.241 & 0.226 & 0.258 \\
Eligible women $\times$ education & 1.244 & 1.188 & 1.303 \\
Eligible women $\times$ wealth & 0.806 & 0.757 & 0.859 \\
Wealth $\times$ education & 1.235 & 1.170 & 1.304 \\
Urban $\times$ wealth & 0.900 & 0.850 & 0.954 \\
Eligible women $\times$ urban & 1.084 & 1.041 & 1.128 \\
Urban $\times$ education & 1.044 & 1.008 & 1.082 \\
Age $\times$ education & 1.044 & 0.988 & 1.103 \\
\bottomrule
\end{tabular}
\end{table}

Seven of the ten estimable interaction terms reach the conventional 5\% significance level. Given the large sample size, the analysis emphasizes effect size and out-of-sample contribution rather than statistical significance alone.

\subsection{Held-Out Interaction Test}
Adding explicit interaction products to the additive logistic specification increases held-out ROC-AUC by 0.0027 and PR-AUC by 0.0008. In the nested feature-block analysis, the interaction block changes ROC-AUC by $-0.0002$ for HistGradientBoosting, $+0.0001$ for logistic regression, and $+0.0005$ for LightGBM.

Thus, explicit interactions contain detectable structure but provide little additional predictive discrimination once the underlying variables and flexible nonlinear relationships are already represented.

\begin{figure}[H]
\centering
\includegraphics[width=\columnwidth]{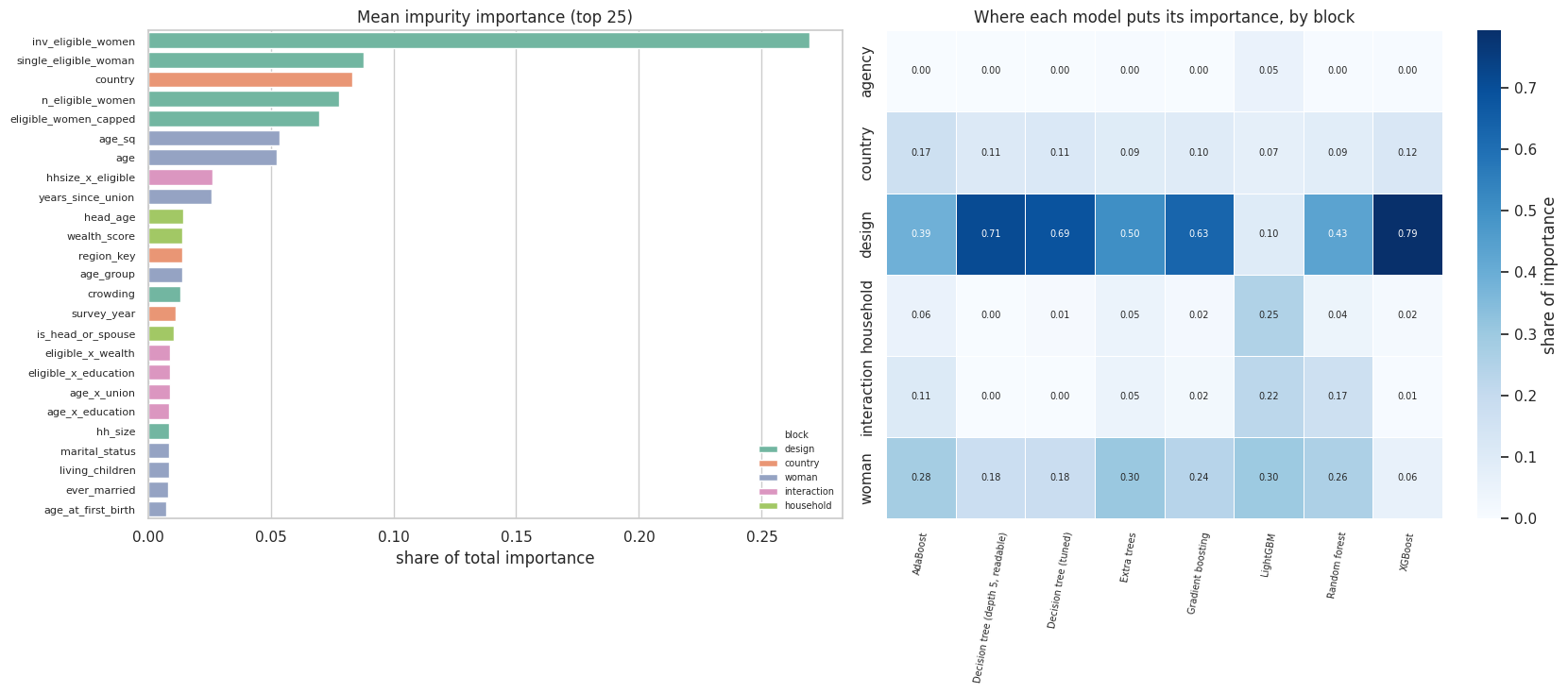}
\caption{Feature importance and allocation of importance across feature blocks. The left panel ranks the top individual features; the right panel aggregates importance by model block.}
\label{fig:app_feature_importance}
\end{figure}

\begin{figure}[H]
\centering
\includegraphics[width=1\columnwidth]{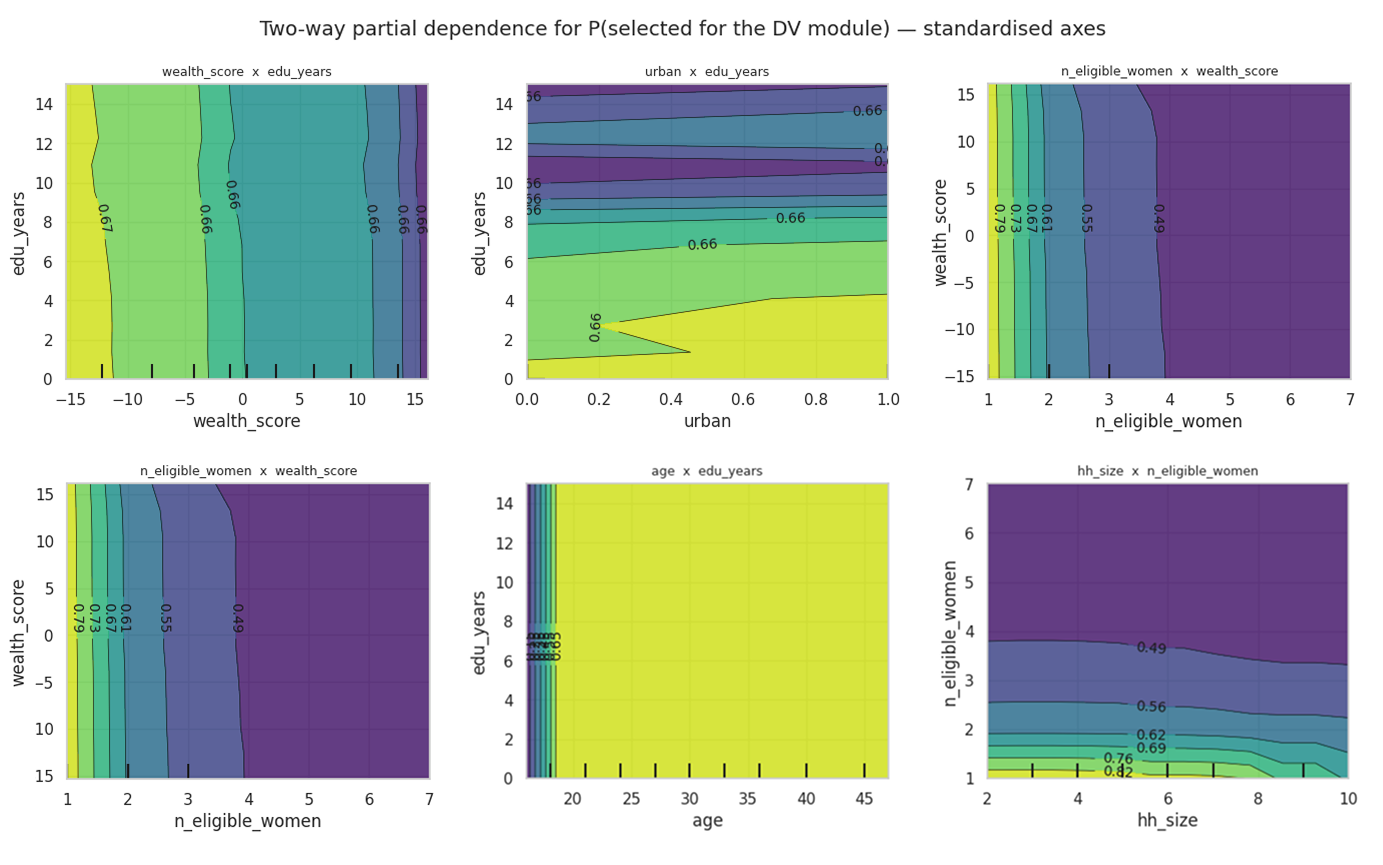}
\caption{Two-way partial dependence for the predicted probability of DV-module selection. Joint surfaces illustrate nonlinear relationships between selected feature pairs and the modelled probability of selection.}
\label{fig:app_two_way_pdp}
\end{figure}

\begin{figure}[H]
\centering
\includegraphics[width=1\columnwidth]{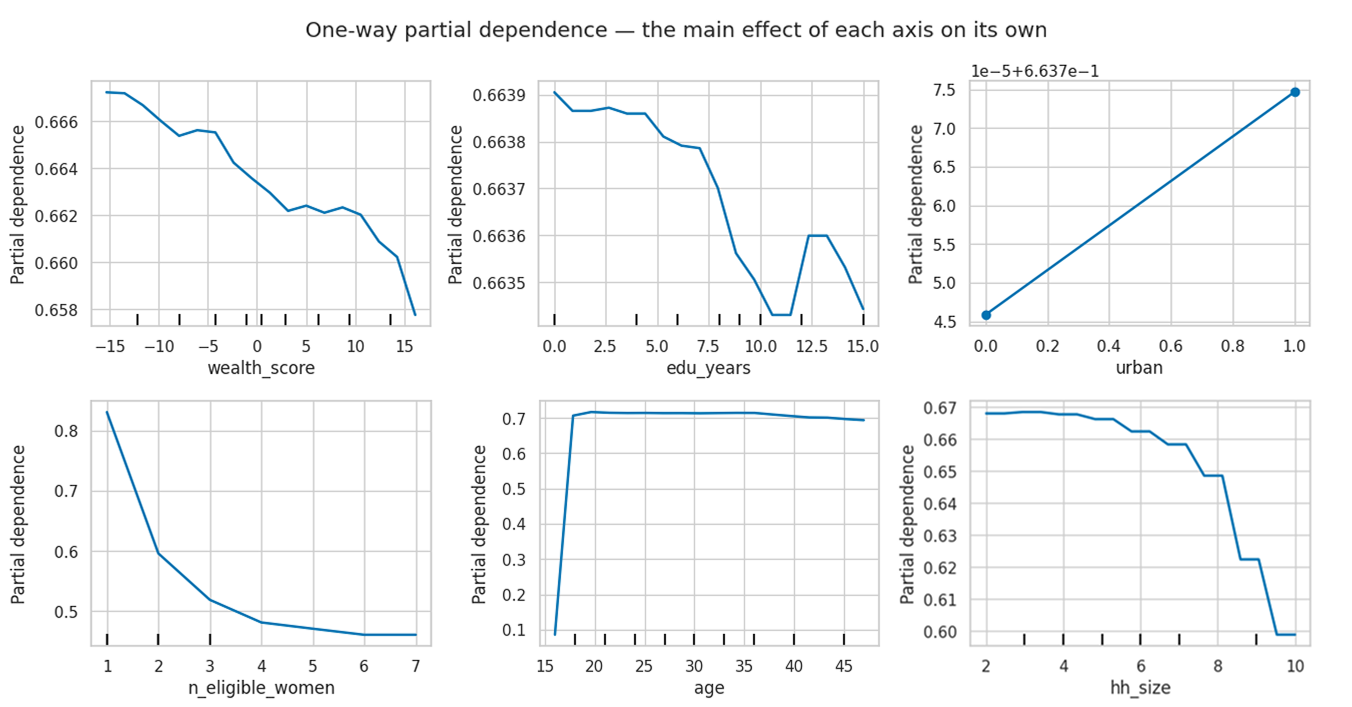}
\caption{One-way partial dependence for the predicted probability of DV-module selection. Curves summarize the modelled marginal relationship between individual feature dimensions and predicted selection probability.}
\label{fig:app_one_way_pdp}
\end{figure}

Figure~\ref{fig:app_feature_importance} ranks the most important features and aggregates importance across feature blocks. Figures~\ref{fig:app_two_way_pdp} and~\ref{fig:app_one_way_pdp} provide complementary joint and marginal partial-dependence visualizations.

\section{Cross-Country Analysis}
\label{app:cross_country}

The pooled analysis is complemented by country-specific and leave-one-country-out evaluations to assess the stability of the learned profile-selection relationship across national contexts. Table~\ref{tab:appendix_country_models} summarizes within-country performance relative to the country-specific design baseline, while Table~\ref{tab:appendix_loco} reports transfer performance when each country is withheld from model training.

\subsection{Within-Country Performance}

The country-specific models use cluster-aware holdout evaluation within each country. Table~\ref{tab:appendix_country_models} reports the profile-model ROC-AUC, the country-specific design-only ROC-AUC, and the corresponding gain.

\begin{table}[H]
\centering
\caption{Within-country profile-model performance relative to design-only baselines.}
\label{tab:appendix_country_models}
\small
\begin{tabular}{lrrr}
\toprule
Country & Profile model ROC-AUC & Design ROC-AUC & Gain \\
\midrule
India & 0.9196 & 0.8486 & +0.0710 \\
Afghanistan & 0.8713 & 0.8746 & -0.0033 \\
Pakistan & 0.6102 & 0.6268 & -0.0166 \\
Nepal & 0.6331 & 0.6642 & -0.0311 \\
Myanmar & 0.6427 & 0.6771 & -0.0344 \\
\bottomrule
\end{tabular}
\end{table}

The profile gain therefore ranges from $+0.0710$ ROC-AUC in India to $-0.0344$ in Myanmar. Country-specific feature-importance rankings remain moderately to strongly aligned, with a mean off-diagonal Spearman correlation of 0.791 and pairwise correlations ranging from 0.578 to 0.925.

\begin{figure}[H]
\centering
\includegraphics[width=\columnwidth]{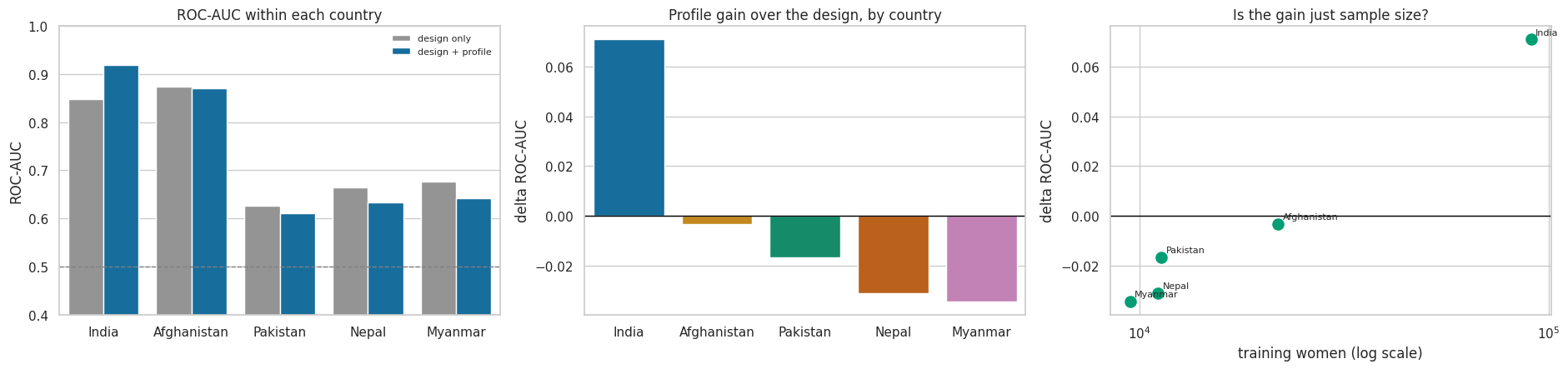}
\caption{Within-country ROC-AUC, profile gain over the design baseline, and the relationship between profile gain and sample size.}
\label{fig:app_country_within}
\end{figure}

\subsection{Leave-One-Country-Out Transfer}
For each country, the transfer model is trained on the remaining countries and evaluated on the held-out country. The held-out country's indicator is excluded from the transfer feature space so that the model cannot exploit an artificial target-country category.

Table~\ref{tab:appendix_loco} reports the resulting transfer ROC-AUC and summarizes the mean transfer penalty.

\begin{table}[H]
\centering
\caption{Leave-one-country-out transfer performance.}
\label{tab:appendix_loco}
\small
\begin{tabular}{lr}
\toprule
Held-out country & Transfer ROC-AUC \\
\midrule
Afghanistan & 0.8719 \\
India & 0.8374 \\
Myanmar & 0.6668 \\
Nepal & 0.6484 \\
Pakistan & 0.6219 \\
\midrule
Mean transfer penalty & -0.0061 \\
\bottomrule
\end{tabular}
\end{table}

The mean transfer penalty is $-0.0061$ ROC-AUC, indicating that a substantial portion of the broad predictive structure transfers across countries. However, the mean transferred gain relative to the country-specific design baseline is $-0.0076$, indicating that the incremental profile advantage does not consistently survive country shifts.

\begin{figure}[H]
\centering
\includegraphics[width=\columnwidth]{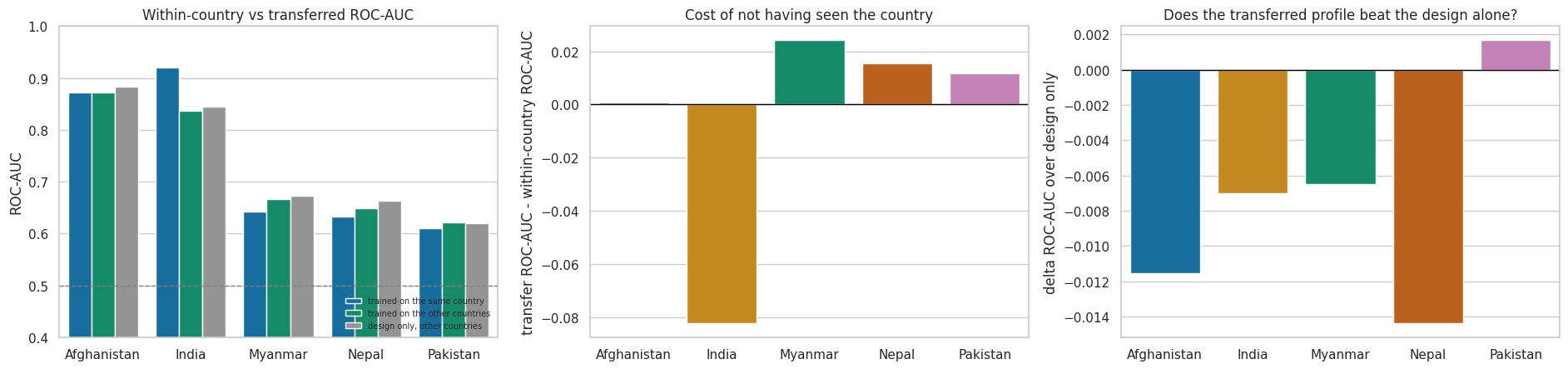}
\caption{Within-country versus leave-one-country-out transfer performance. The panels report absolute transfer performance, transfer penalties, and whether the transferred profile model exceeds the target-country design baseline.}
\label{fig:app_country_transfer}
\end{figure}

\end{document}